\pdfoutput=1
\documentclass{article}
\usepackage{iclr2027_conference,times}
\iclrpreprintcopy
\usepackage[utf8]{inputenc}
\usepackage[T1]{fontenc}
\usepackage{amsmath,amssymb,mathtools,graphicx,booktabs,array,tabularx,longtable}
\usepackage{microtype}
\usepackage{xcolor,tikz}
\usepackage{fontawesome5}
\usetikzlibrary{arrows.meta,positioning,fit}
\usepackage[hidelinks]{hyperref}
\usepackage{url}
\newcommand{\method}{Concertina}
\newcommand{\evlm}{eVLM}
\DeclareRobustCommand{\cmark}{$\checkmark$}
\DeclareRobustCommand{\xmark}{$\times$}
\DeclareRobustCommand{\hmark}{\tikz[baseline=-0.55ex]{\draw[line width=0.45pt] (0,0) circle (0.62ex); \fill (0,0.62ex) arc (90:270:0.62ex) -- cycle;}}
\AtBeginDocument{%
  \setlength{\abovedisplayskip}{4.5pt plus 1.5pt minus 1.5pt}%
  \setlength{\belowdisplayskip}{4.5pt plus 1.5pt minus 1.5pt}%
  \setlength{\abovedisplayshortskip}{2pt plus 1pt minus 1pt}%
  \setlength{\belowdisplayshortskip}{4.5pt plus 1.5pt minus 1.5pt}%
  \setlength{\textfloatsep}{10pt plus 2pt minus 2pt}%
  \setlength{\floatsep}{8pt plus 2pt minus 2pt}%
}
\title{\centering\makebox[\textwidth][c]{\fontsize{16.2pt}{20.5pt}\selectfont\raisebox{-0.30\height}{\includegraphics[height=1.62em]{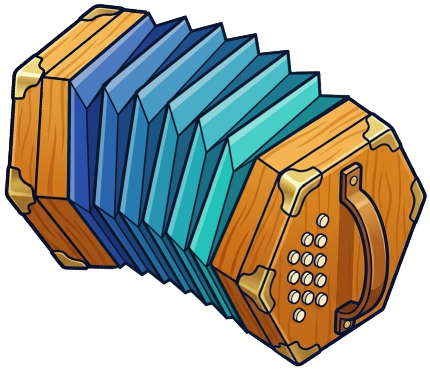}}\hspace{0.34em}\method: Embodying Manipulation and Cognition}\\in One Diffusion Vision-Language Model}
\makeatletter
\author{%
\kern-\tabcolsep
\parbox{\textwidth}{\centering
\textbf{Chengcheng Wang$^{1}$\hspace{2.0em}Wei He$^{2}$\hspace{2.0em}Wenhao Li$^{1}$\hspace{2.0em}Hongguang Li$^{2}$} \\[2.5pt]
\textbf{Kai Han$^{2}$\hspace{2.0em}Jianyuan Guo$^{3}$\hspace{2.0em}Chang Xu$^{1,\dagger}$} \\[5pt]
\normalfont $^{1}$University of Sydney\hspace{1.6em}$^{2}$TokenRhythm Technologies\hspace{1.6em}$^{3}$City University of Hong Kong \\[3.5pt]
{\normalfont\footnotesize\texttt{\{cwan0785, weli0055\}@uni.sydney.edu.au, c.xu@sydney.edu.au}} \\[0.5pt]
{\normalfont\footnotesize\texttt{\{wei.he, hongguang.li, kai.han\}@tokenrhythm.ai, jianyguo@cityu.edu.hk}} \\[8pt]
\makebox[\textwidth][c]{\normalfont\small
\href{https://concertina.traly.org/}{\raisebox{-0.12ex}{\textcolor[HTML]{1A569E}{\faGlobe}}\hspace{0.38em}\textcolor[HTML]{1A569E}{\textbf{Homepage}}}\hspace{2.2em}%
\href{https://github.com/lose4578/Concertina}{\raisebox{-0.12ex}{\faGithub}\hspace{0.38em}\textcolor[HTML]{1A569E}{\textbf{GitHub}}}\hspace{2.2em}%
\href{https://huggingface.co/collections/Traly/concertina}{\raisebox{-0.22ex}{\includegraphics[height=1.18em]{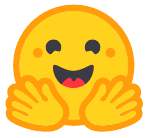}}\hspace{0.35em}\textcolor[HTML]{1A569E}{\textbf{Hugging Face}}}%
}\vspace{-1.30em}
}\kern-\tabcolsep%
\gdef\@thanks{\footnotetext[2]{Corresponding author.}}%
}
\makeatother
\begin{document}
\maketitle
\begin{abstract}
Most vision-language-action (VLA) policies decouple manipulation from cognition by bolting an external action mechanism onto a vision-language model (VLM)---either quantizing actions into discrete tokens whose errors bear no stable correspondence to physical displacements, or delegating continuous trajectory generation to a separate action expert whose gradients compete with or are blocked from updating the VLM backbone. We argue that manipulation and cognition should instead be embodied in the same model, shifting from a decoupled VLA to an \emph{Embodied Vision-Language Model} (\textbf{\evlm}). We introduce \method, an \evlm{} that places continuous trajectory representation, blockwise generation, and physical-coordinate supervision directly on the computation and parameter path of a shared diffusion VLM backbone. Each control timestep is encoded as one continuous latent action token in the backbone hidden space, and the shared Transformer reconstructs temporally ordered action blocks via the same corrupt-and-reconstruct operator used for multimodal pretraining. Active action projections account for only $0.21\%$ of parameters in the 4.46B model with zero separate action-generator layers, compared with $16.5\%$--$35.1\%$ dedicated action-expert capacity in representative continuous VLAs. Using our proposed \emph{Action Interface Audit}, we confirm that cross-time trajectory coupling emerges in the middle and late layers of the shared backbone rather than at the action interfaces. On standard benchmarks, \method{} achieves $98.15\%$ average success on regular LIBERO, $85.9\%$ on the 10{,}030-task LIBERO-Plus benchmark, and $4.142$ average completed chain length on CALVIN ABC--D.
\end{abstract}
\section{Introduction}
\label{sec:introduction}
The premise of vision-language-action (VLA) policies is to transfer the visual perception, semantic grounding, and generalization of pretrained vision-language models (VLMs) to physical manipulation. Beneath specific design choices lies a foundational question: \emph{how can we enable a pretrained VLM to natively output precise continuous actions while preserving its visual-language generalization, its model architecture, and its native training paradigm?} Existing VLAs largely compromise on at least one of these requirements by treating action as an external attachment to a VLM---dividing manipulation and cognition across separate representations, modules, or training objectives, where the VLM handles scene and instruction understanding while another mechanism handles action representation, trajectory generation, or coordinate supervision. Our thesis is that manipulation and cognition should not be decoupled across a bolted-on VLA pipeline; they should be embodied in a single \emph{Embodied Vision-Language Model} (\textbf{\evlm}). Concretely, in an \evlm, actions are represented directly in the backbone's continuous hidden space, generated across time by the same parameters and native generative operator that process vision and language, and supervised in physical coordinates end-to-end without degrading the shared VLM.

Three principles motivate this shift from decoupled VLAs to a unified \evlm. First, \emph{cognition should participate in manipulation at every layer.} Grasping, placement, and obstacle avoidance rely on the same spatial and physical cues as scene understanding; when trajectory generation happens outside the VLM, cognition is reduced to a one-time conditioning handoff and cannot interact with evolving action states across depth. Second, \emph{manipulation should shape cognition in return.} Robot trajectories are the primary source of physical and spatial grounding for a VLM; only when continuous action gradients directly update the vision-language parameters can manipulation experience consolidate into the backbone's spatial cognition rather than remaining confined to an external head. Third, \emph{preserving the backbone's architecture, training paradigm, and representation stability enables end-to-end scaling.} When cognitive and motor modules rely on incompatible paradigms or objectives, their gradients compete and degrade pretrained representations, often forcing practitioners to insulate or freeze the backbone~\citep{knowledgeinsulation,ddvla} and permanently fixing the separation between perception and control.

From this perspective, conventional VLAs fail to form a unified \evlm{} in two primary ways (Table~\ref{tab:arch-properties}). The first is \emph{representational decoupling}. Methods such as RT-2~\citep{rt2} and OpenVLA~\citep{openvla} quantize each action dimension into discrete bins and reuse the language vocabulary under cross-entropy loss, while FAST~\citep{pi0fast} compresses action chunks via a discrete cosine transform and byte-pair encoding. Although these methods preserve the VLM architecture and training paradigm, control is moved from physical space into symbol space: token-level cross-entropy penalizes adjacent-bin and far-bin errors equally, and compressed tokens entangle multiple timesteps and dimensions so that a single token error can corrupt an entire trajectory. The second is \emph{computational and parameter decoupling}. Continuous policies such as $\pi_0$~\citep{pi0}, $\pi_{0.5}$~\citep{pi05}, SmolVLA~\citep{smolvla}, GR00T N1~\citep{gr00tn1}, and X-VLA~\citep{xvla} preserve physical coordinates via flow matching or diffusion, but alter the model architecture and training paradigm by delegating trajectory generation to a dedicated action expert or 24-layer action Transformer outside the VLM. Because action supervision primarily updates the external expert and its competing gradients can degrade pretrained VLM features, Knowledge Insulation~\citep{knowledgeinsulation} blocks expert gradients from entering the backbone---resolving training instability by hard-wiring architectural decoupling. Recent hybrids partially restore coupling along individual axes---sharing a backbone over discrete diffusion tokens~\citep{ddvla,mmadavla,blockvla,tbdvla}, combining discrete and continuous branches~\citep{hybridvla}, training a separate trajectory codec~\citep{cat}, or switching from discrete pretraining to downstream flow matching~\citep{dreamvla}---yet none realizes a true \evlm{} that natively generates continuous physical trajectories while preserving its model architecture, its native pretraining paradigm, and its visual-language generalization.

\begin{table*}[t]
\centering
\vspace{-0.6em}
\scriptsize
\setlength{\tabcolsep}{3.5pt}
\renewcommand{\arraystretch}{1.12}
\resizebox{\textwidth}{!}{%
\begin{tabular}{@{}>{\raggedright\arraybackslash}p{1.68in} *{10}{c}@{}}
\toprule
& \multicolumn{2}{c}{Representational Coupling} & \multicolumn{4}{c}{Computational \& Parameter Coupling} & \multicolumn{4}{c}{Generation \& Complexity} \\
\cmidrule(lr){2-3}\cmidrule(lr){4-7}\cmidrule(lr){8-11}
Paradigm / Method & \shortstack{Cont.\ action\\values} & \shortstack{Coordinate\\supervision} & \shortstack{Backbone\\generates traj.} & \shortstack{No external\\action gen.} & \shortstack{No pretrained\\tokenizer/codec} & \shortstack{Single action\\repr.\ \& loss} & \shortstack{Parallel / joint\\generation} & \shortstack{Causal\\temporal order} & \shortstack{Iterative\\refinement} & \shortstack{Sequence\\length $\mathcal{O}(H)$} \\
\midrule
RT-2~\citep{rt2} / OpenVLA~\citep{openvla} & \xmark & \xmark & \cmark & \cmark & \cmark & \cmark & \xmark & \cmark & \xmark & \xmark \\
FAST / $\pi_0$-FAST~\citep{pi0fast} & \xmark & \xmark & \cmark & \cmark & \xmark & \cmark & \xmark & \hmark & \xmark & \hmark \\
$\pi_0$~\citep{pi0} / $\pi_{0.5}$~\citep{pi05} & \cmark & \cmark & \xmark & \xmark & \cmark & \cmark & \cmark & \xmark & \cmark & \cmark \\
SmolVLA~\citep{smolvla} / GR00T N1~\citep{gr00tn1} & \cmark & \cmark & \xmark & \xmark & \cmark & \cmark & \cmark & \xmark & \cmark & \cmark \\
X-VLA~\citep{xvla} & \cmark & \cmark & \xmark & \xmark & \cmark & \cmark & \cmark & \xmark & \cmark & \cmark \\
HybridVLA~\citep{hybridvla} & \cmark & \hmark & \cmark & \cmark & \cmark & \xmark & \cmark & \xmark & \cmark & \cmark \\
Dream-VLA~\citep{dreamvla} & \hmark & \hmark & \hmark & \cmark & \cmark & \xmark & \cmark & \xmark & \cmark & \cmark \\
Discrete Diffusion VLA~\citep{ddvla} / MMaDA-VLA~\citep{mmadavla} & \xmark & \xmark & \cmark & \cmark & \cmark & \cmark & \cmark & \xmark & \cmark & \xmark \\
BlockVLA~\citep{blockvla} / TBD-VLA~\citep{tbdvla} & \xmark & \xmark & \cmark & \cmark & \cmark & \cmark & \cmark & \cmark & \cmark & \xmark \\
CAT~\citep{cat} & \cmark & \hmark & \xmark & \xmark & \xmark & \cmark & \cmark & \xmark & \cmark & \cmark \\
\midrule
\textbf{\method{} (\evlm, ours)} & \cmark & \cmark & \cmark & \cmark & \cmark & \cmark & \cmark & \cmark & \cmark & \cmark \\
\bottomrule
\end{tabular}%
}
\vspace{-0.4em}
\caption{Architectural comparison between representative VLA paradigms and our \textbf{\evlm} (\cmark: satisfied; \xmark: not satisfied; \hmark: partially satisfied). Columns 1--2 reflect representational coupling (whether actions enter and are supervised in physical coordinates); columns 3--4 reflect computational and parameter coupling (whether the VLM backbone directly generates trajectories without a separate action Transformer, DiT, flow expert, or post-codec generator); the remaining columns summarize training and decoding structure. Coordinate supervision marks \hmark{} for dual discrete-continuous objectives (HybridVLA), latent-codec policy losses (CAT), or discrete-pretrain/continuous-finetune stage switches (Dream-VLA).}
\vspace{-0.6em}
\label{tab:arch-properties}
\end{table*}

We instantiate this \evlm{} paradigm in \textbf{\method} through an end-to-end \emph{continuous latent action representation} inside a diffusion VLM. Prior latent-action works (LAPA, IGOR, UniVLA, Moto, villa-X, CAT~\citep{lapa,igor,univla_latent,moto,villax,cat}) show that mapping diverse robot controls into a shared latent space decouples embodiment-specific kinematics from sequence modeling, but they rely on separately pretrained VQ codebooks or trajectory codecs. We place this latent space directly inside the hidden space of a diffusion VLM (BARD-VL~\citep{bard}): at each control timestep, a lightweight position-wise action encoder maps the complete $D$-dimensional action vector into a single continuous latent action token in $\mathbb{R}^{d}$, so a horizon-$H$ trajectory occupies $H$ tokens rather than $HD$ scalar symbols. Vision, language, and latent action tokens are jointly contextualized across every layer of the shared Transformer, and a lightweight position-wise action decoder maps output hidden states back to physical coordinates under direct coordinate reconstruction loss.

\begin{figure*}[t]
\centering
\vspace{-0.4em}
\includegraphics[width=0.92\textwidth]{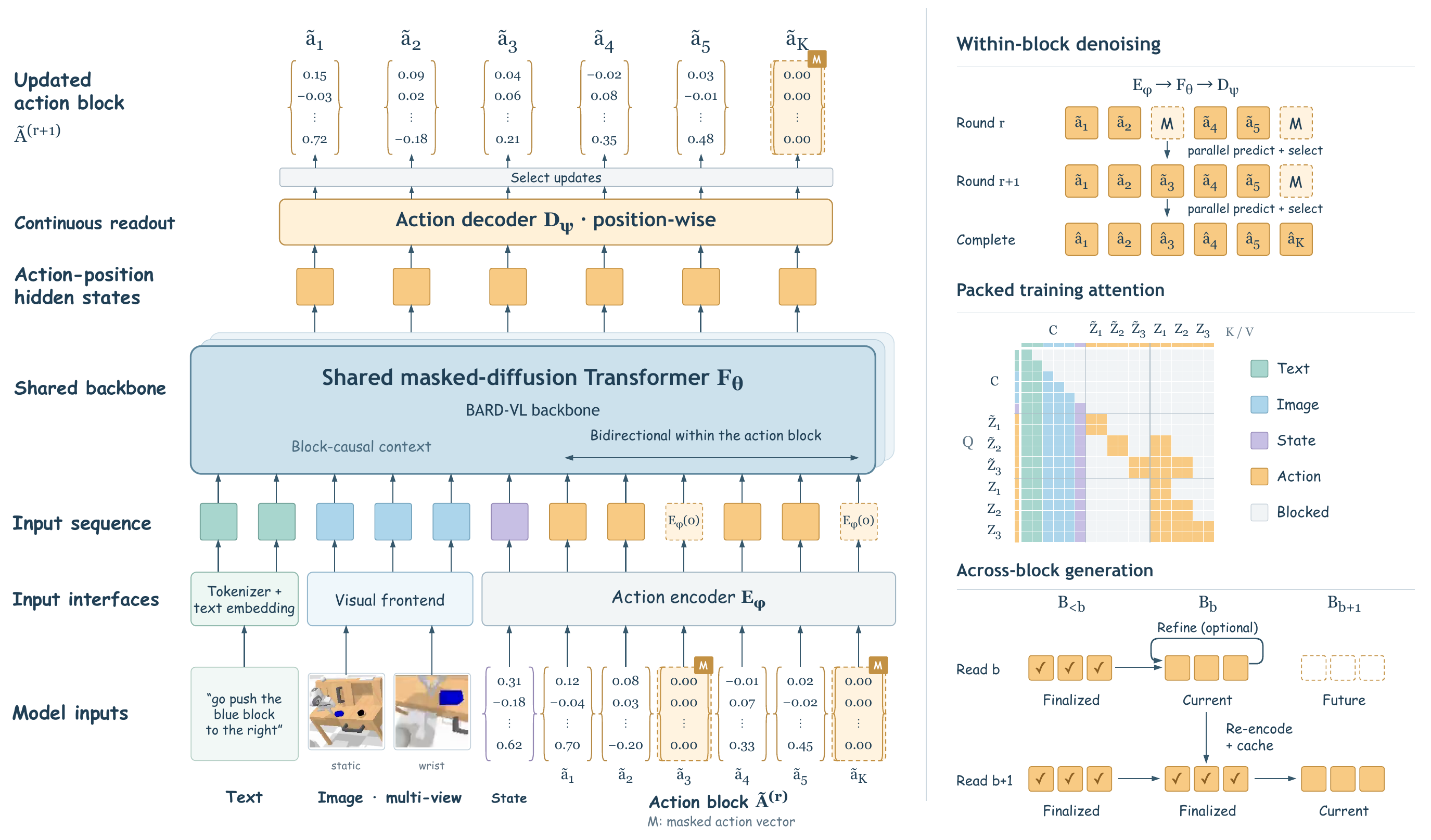}
\vspace{-0.6em}
\caption{\textbf{\method{} (\evlm) reconstructs continuous action blocks inside a shared diffusion vision-language model.} Multimodal inputs and continuous latent action tokens are jointly processed across all layers of the shared Transformer $F_\theta$, combining position-wise continuous coordinate projections ($E_\phi, D_\psi$), packed blockwise teacher forcing during training, and risk-guided blockwise reconstruction at inference. Amber tiles denote complete action vectors; dashed ``M'' tiles denote masked vectors.}
\vspace{-0.4em}
\label{fig:overview}
\end{figure*}

Building on this principle, \method{} partitions the $H$-step horizon into temporally ordered action blocks that are causal across blocks and bidirectional within each block, allowing contiguous action segments to be generated and refined jointly. During training, the multimodal conditioning prefix, corrupted action tokens, and teacher-forced clean action tokens are packed into a single sequence, supervising all block-conditional distributions in one forward pass and providing a compute-efficient, scalable end-to-end training path. Because the VLM backbone is already pretrained with masked block reconstruction~\citep{bard}, \method{} reuses the exact same corrupt-and-reconstruct generative operator for continuous control without introducing a second training paradigm or external diffusion network.

To rigorously test whether a policy acts as a decoupled VLA or a true \evlm---and to localize where action representation and trajectory computation reside---we introduce the \textbf{Action Interface Audit}, a checkpoint-level methodology comprising three metrics aligned with the three decoupling axes: (i) \emph{Error-Geometry Consistency} ($\mathrm{EGC}=p_{95}/p_{05}$), measuring whether unit training errors map stably to physical action errors; (ii) \emph{Trajectory Coupling Attribution} (TCA) via the non-local response ratio ($\mathrm{NLR}$), using deterministic finite differences to localize where cross-time dependencies emerge; and (iii) \emph{Action-Specific Capacity Ratio} ($\mathrm{ASCR}$), quantifying parameters dedicated solely to action generation. The audit reveals that uniform discrete bins induce a $25.14\times$ physical-error span per single-token error and FAST-style compressed tokens suffer $77.9\%$--$82.2\%$ structural decoding failures under single-token edits; $\pi_0$ and $\pi_{0.5}$ derive all cross-time coupling from external action experts ($\mathrm{ASCR}=16.51\%$ and $19.17\%$, rising to $22.19\%$ in SmolVLA and $35.10\%$ in X-VLA); whereas \method's active action interfaces occupy only $0.21\%$ of the 4.46B model, exhibit zero cross-time response on their own ($\mathrm{NLR}=0$), and delegate all temporal trajectory coupling to the middle and late layers of the shared VLM backbone ($\mathrm{NLR}=4.06\%$ in pretraining, $4.10\%$ on CALVIN, and $1.22\%$ on LIBERO). Furthermore, a matched counterfactual intervention shows that attaching a second coordinate objective to the shared hidden space induces opposing gradients (cosine $-0.891$) and collapses CALVIN chain length from $2.970$ to $0.237$, confirming that enabling native action generation in an \evlm{} without degrading the shared VLM requires co-designing the action representation and its supervision objective as a unified whole. On standard closed-loop benchmarks, \method{} reaches $98.15\%$ average success on regular LIBERO, $85.9\%$ on the 10{,}030-task LIBERO-Plus benchmark, and $4.142$ average completed chain length on CALVIN ABC--D (Table~\ref{tab:main-libero}).

Our contributions are threefold:
\begin{itemize}
\item \textbf{Embodied Vision-Language Model (\evlm) Paradigm.} We formulate the \evlm{} paradigm, embodying manipulation and cognition within a single VLM across representation, supervision, and generation so that a pretrained VLM natively generates continuous action trajectories while preserving its model architecture, native training paradigm, and visual-language generalization.
\item \textbf{Continuous Blockwise \evlm{} Architecture (\method).} We introduce \method, representing a horizon-$H$ trajectory as $H$ continuous latent action tokens with blockwise causal-bidirectional visibility and packed teacher forcing, requiring only $0.21\%$ active interface parameters and no external action expert, tokenizer, or codec.
\item \textbf{Action Interface Audit and Empirical Validation.} We introduce EGC, TCA ($\mathrm{NLR}$), and ASCR to audit representative discrete VLAs, action-expert VLAs, and our \evlm, provide counterfactual evidence on shared-space objective design, and demonstrate state-of-the-art closed-loop performance on LIBERO and LIBERO-Plus alongside strong zero-shot transfer on CALVIN ABC--D.
\end{itemize}

\section{Related Work}
\label{sec:related_work}

\paragraph{Latent action representations.}
Several recent works learn compact latent action variables from video or robot demonstrations before policy learning. LAPA and IGOR extract discrete latent actions from action-free videos and image-goal pairs before grounding them into physical controls in a subsequent stage~\citep{lapa,igor}, while UniVLA~\citep{univla_latent}, Moto~\citep{moto}, villa-X~\citep{villax}, and CAT~\citep{cat} introduce task-centric codebooks, motion tokens, proprioceptive latent dynamics, or frozen trajectory codecs. Because these methods freeze the tokenizer or codec prior to policy optimization, the latent space is never shaped directly by closed-loop physical control gradients. By contrast, \method{} embeds continuous actions directly into the hidden space of the diffusion VLM through lightweight position-wise projections co-trained end-to-end with the backbone under physical-coordinate reconstruction loss.

\paragraph{External action experts versus shared-backbone generation.}
Building on diffusion and flow-matching policies~\citep{diffusionpolicy}, continuous VLA architectures such as $\pi_0$~\citep{pi0}, $\pi_{0.5}$~\citep{pi05}, SmolVLA~\citep{smolvla}, GR00T N1~\citep{gr00tn1}, and X-VLA~\citep{xvla} preserve continuous control coordinates by pairing a VLM with a dedicated flow-matching action expert or a 24-layer external action Transformer, while OpenVLA-OFT attaches a continuous regression head~\citep{openvlaoft}. Delegating trajectory generation to an external module separates motor computation from visual-language reasoning and introduces cross-module gradient tension; Knowledge Insulation~\citep{knowledgeinsulation} addresses the resulting backbone degradation by blocking expert gradients from updating the VLM, permanently fixing the division between perception and control. \method{} eliminates the external action generator altogether, executing trajectory reconstruction inside the shared VLM Transformer so that manipulation and cognition co-adapt along a single parameter path.

\paragraph{Discrete and compressed action tokenizers.}
A third paradigm represents robot controls as discrete symbols over the language vocabulary. RT-2 and OpenVLA quantize each action coordinate into uniform bins~\citep{rt2,openvla}; FAST compresses action chunks via a discrete cosine transform and byte-pair encoding~\citep{pi0fast}; OAT learns ordered discrete action tokens~\citep{oat}; and Discrete Diffusion VLA, MMaDA-VLA, and LLaDA-VLA apply masked discrete diffusion over quantized action tokens~\citep{ddvla,mmadavla,lladavla}. Operating in symbol space decouples the training loss from physical error geometry: cross-entropy penalizes near-miss and large-displacement bin errors equally, while frequency-compressed tokens entangle multiple timesteps so that a single token error can invalidate an entire action chunk. \method{} retains token-level sequence length $\mathcal{O}(H)$ while supervising every token directly in normalized physical coordinates.

\begin{figure*}[t]
\centering
\vspace{-0.4em}
\includegraphics[width=0.96\textwidth]{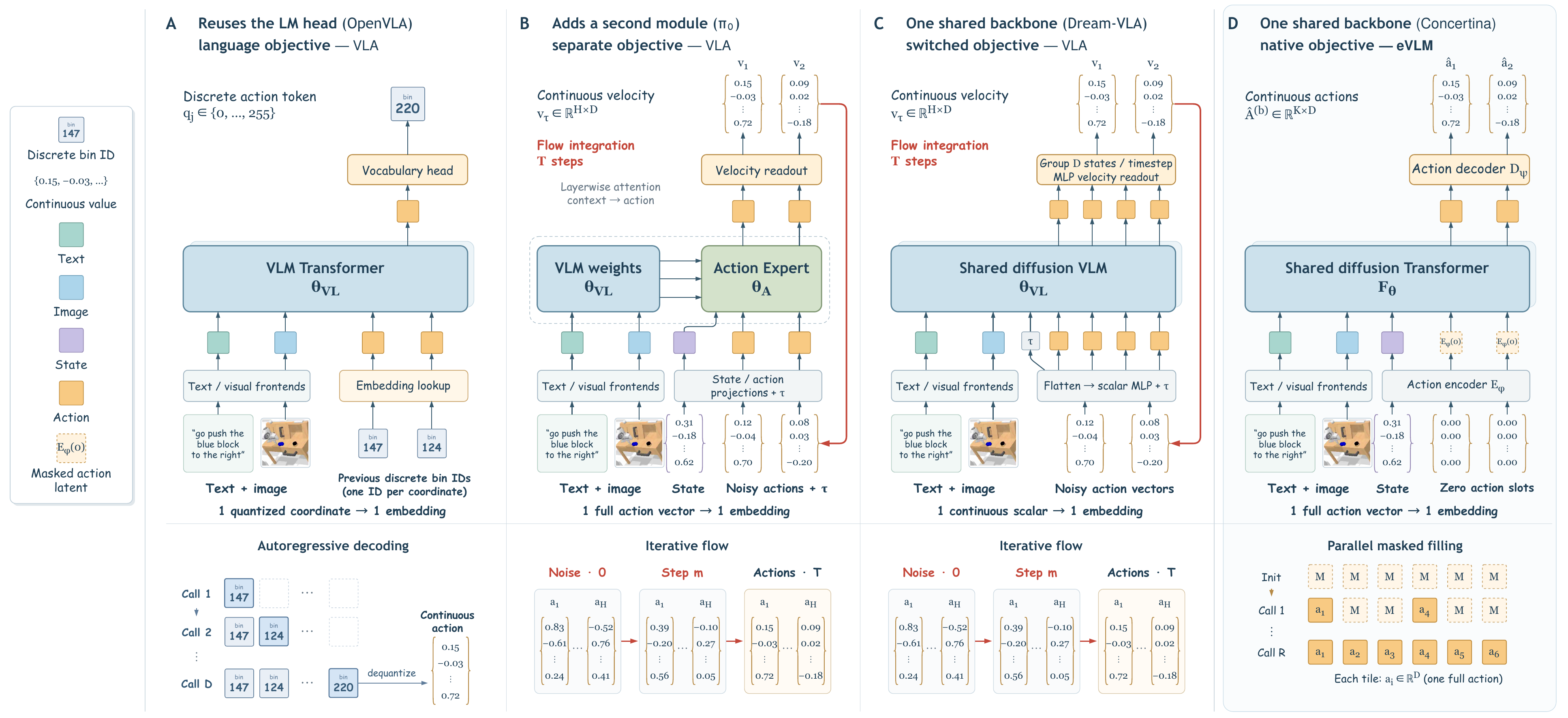}
\vspace{-0.8em}
\caption{\textbf{From decoupled VLAs to an Embodied Vision-Language Model (\evlm).} Comparison of four architectural routes to robot manipulation: (A) OpenVLA quantizes coordinates into discrete vocabulary tokens (language-objective VLA); (B) $\pi_0$ delegates continuous trajectory generation to a separate flow-matching action expert (separate-objective VLA); (C) Dream-VLA retains a shared diffusion VLM backbone but switches to a flow-matching objective for continuous control (switched-objective VLA); (D) \method{} instantiates an \textbf{\evlm} that preserves both the shared diffusion VLM backbone and its native masked block reconstruction operator over continuous latent action tokens.}
\vspace{-0.4em}
\label{fig:architecture-comparison}
\end{figure*}

\paragraph{Blockwise diffusion VLMs.}
Blockwise diffusion models combine causal ordering across blocks with parallel bidirectional denoising within each block. BARD-VL bridges autoregressive and diffusion vision-language modeling via progressive block merging and stage-wise distillation over packed clean and corrupted sequences~\citep{bard}. Recent VLA adaptations apply block diffusion to discretized action bins~\citep{blockvla,tbdvla}, combine autoregressive and continuous diffusion branches~\citep{hybridvla}, or switch from discrete masked pretraining to downstream flow matching~\citep{dreamvla}. \method{} instead generalizes masked block reconstruction directly to continuous latent action vectors, turning the diffusion VLM itself into an \evlm{} that preserves both its shared backbone and its native pretraining operator (Figure~\ref{fig:architecture-comparison}).

\section{Method}
\label{sec:method}

\method{} instantiates an \emph{Embodied Vision-Language Model} (\evlm) that embodies manipulation and cognition within a single diffusion vision-language model, enabling the pretrained backbone to natively generate continuous action trajectories without altering its model architecture or training paradigm. Its atomic generation unit is one complete continuous action vector at one control timestep, represented as a continuous latent action token: a lightweight position-wise encoder maps it into the VLM hidden space, and a position-wise decoder maps the contextualized hidden state back to physical control coordinates. All cross-time and cross-modal dependencies are modeled by the shared Transformer $F_\theta$ without an external action expert or discrete action tokenizer. This design establishes two direct coupling pathways between cognition and control: in the \emph{forward pass}, action tokens attend to visual and linguistic representations at every Transformer layer rather than receiving a one-time conditioning summary; in the \emph{backward pass}, physical-coordinate reconstruction gradients update the shared vision-language parameters end-to-end, consolidating manipulation experience into the backbone's spatial representations.

\subsection{Preliminaries and Continuous Latent Action Tokens}
\label{sec:method-preliminaries}

Given visual observations $\mathcal I$, a language instruction $l$, and proprioceptive state $s$, the policy predicts a horizon-$H$ continuous action trajectory $A=(a_1,\ldots,a_H)\in\mathbb R^{H\times D}$, where $a_i\in\mathbb R^D$. The VLM encodes visual, language, and state inputs into a multimodal conditioning prefix $C=(c_1,\ldots,c_P)$. \method{} initializes $F_\theta$ from BARD-VL~\citep{bard}, reusing its masked blockwise corrupt-and-reconstruct operator across continuous latent action tokens so that cognitive pretraining and motor learning share the same generative procedure.

A position-wise action encoder $E_{\phi}$ and action decoder $D_{\psi}$ bridge physical control coordinates and the $d$-dimensional backbone hidden space:
\begin{equation}
z_i=E_{\phi}(a_i)\in\mathbb R^d,
\qquad \widehat a_i=D_{\psi}(h_i)\in\mathbb R^D.
\label{eq:action-interface}
\end{equation}
Both $E_{\phi}$ and $D_{\psi}$ are lightweight multilayer perceptrons applied independently at each timestep $i$ and trained jointly with $F_\theta$. Because they act strictly position-wise, they perform coordinate projection without temporal mixing; all trajectory-level dependencies across the $H$ latent action tokens $Z=(z_1,\ldots,z_H)$ are computed inside $F_\theta$:
\begin{equation}
[H_C,H_A]=F_{\theta}([C,Z]),\qquad \widehat A=D_{\psi}(H_A).
\end{equation}
Representing the full vector $a_i$ as a single token preserves the physical coupling among translation, rotation, and gripper coordinates within each control step while keeping the action sequence length to $H$ tokens; local Jacobian analysis confirms that $E_\phi$ maintains a non-collapsed, full-rank $D$-dimensional chart inside $\mathbb R^d$ throughout pretraining (effective rank $18.63/20$; Section~\ref{sec:temporal-response}).

\paragraph{Unified latent space across embodiments.}
For multi-embodiment training, heterogeneous control spaces are mapped into a shared maximum coordinate width $D$. A binary dimension-validity mask $q_{n,d}\in\{0,1\}$ indicates active coordinates for sample $n$, and a timestep-validity mask $v_{n,i}\in\{0,1\}$ indicates valid trajectory positions. Inactive coordinates are zeroed at the input and excluded from the loss denominator, allowing diverse robot embodiments to share a unified $d$-dimensional latent space inside $F_\theta$ while preserving exact physical supervision on each embodiment's active degrees of freedom.

\subsection{Blockwise Topology and Packed Teacher Forcing}
\label{sec:method-packing}

We partition the horizon $H$ into $M=\lceil H/K\rceil$ temporally ordered action blocks of size at most $K$, with block index $b(i)=\lfloor(i-1)/K\rfloor$ and $A=(A^{(0)},\ldots,A^{(M-1)})$. Generation is causal across blocks and bidirectional within each block: all tokens in block $b$ are reconstructed jointly conditioned on the multimodal prefix $C$, completed earlier blocks $A^{(<b)}$, and the corrupted state $\widetilde A^{(b)}$ of the current block. Inter-block causality preserves the temporal directionality of execution, while intra-block bidirectionality allows contiguous control steps (e.g., approaching, closing the gripper, and lifting) to coordinate and update in parallel as a coherent motion segment.

During training, \method{} packs the multimodal prefix, a corrupted action copy, and a clean teacher-forced action copy into a single Transformer sequence:
\begin{equation}
X_{\mathrm{train}}=[C,\widetilde Z,Z],\qquad \widetilde Z=E_{\phi}(\widetilde A),\quad Z=E_{\phi}(A).
\end{equation}
Both copies share aligned action position identifiers but follow distinct blockwise visibility rules. A corrupted query at timestep $i$ attends to the prefix, its own corrupted block, and strictly earlier clean blocks:
\begin{equation}
\mathcal V_{\mathrm{noisy}}(i)=C\cup\{\widetilde z_j:b(j)=b(i)\}\cup\{z_j:b(j)<b(i)\},
\end{equation}
whereas a clean query attends to the prefix and clean blocks up to its own block, $\mathcal V_{\mathrm{clean}}(i)=C\cup\{z_j:b(j)\leq b(i)\}$. Action predictions are read exclusively from the corrupted branch $\widetilde Z$, while the clean branch $Z$ supplies teacher-forced context to later blocks without leaking current-block targets. Prefix tokens use causal attention within $C$ and do not attend to action branches, so action queries read multimodal features at every layer during the forward pass while action gradients update the shared prefix-processing parameters during backpropagation. This packed layout supervises all $M$ block-conditional distributions simultaneously in a single backbone forward pass over $P+2H$ positions, with sequence length scaling linearly in $H$, providing a compute-efficient and scalable organization for end-to-end pretraining.

\subsection{Block-Stratified Corruption and Coordinate Reconstruction}
\label{sec:method-objective}

Each training sample draws a stratified corruption level for each block so that every forward pass spans the full corruption spectrum. Specifically, we partition $[0,1]$ into $M$ equal strata, sample $u_{n,j}\sim\mathcal U(0,1)$ to form $\bar t_{n,j}=(j+u_{n,j})/M$ for $j=0,\ldots,M-1$, and assign them to blocks via a random permutation $\pi_n$ with lower clipping at $\epsilon$:
\begin{equation}
t_{n,\pi_n(j)}=\operatorname{clip}(\bar t_{n,j},\epsilon,1).
\end{equation}
Within block $b$, each timestep is independently selected for masking according to $m_{n,i}\sim\operatorname{Bernoulli}(t_{n,b(i)})$. Selected positions replace $a_{n,i}$ with an unknown-action sentinel vector (instantiated as the zero vector in $\mathbb R^D$) before encoding through $E_\phi$, while unselected positions retain their clean action vectors to provide intra-block context.\footnote{Appendix~\ref{app:method-details} details both mask-only corruption and the three-state mixed corruption schedule inherited from BARD.}

Supervision is defined directly in normalized physical action coordinates over masked and valid positions. With physical coordinate weights $\omega_d$ ($500:10:1$ across translation, 6D rotation, and gripper coordinates) and coordinate loss $\rho_d$---instantiated as Smooth-$L_1$ regression on continuous pose coordinates and probability-valued binary cross-entropy on the gripper closure coordinate (Appendix~\ref{app:method-details})---the per-token coordinate reconstruction error is
\begin{equation}
\ell_{n,i}
=\frac{\sum_{d=1}^D q_{n,d}\omega_d\,
\rho_d(\widehat a_{n,i,d},\,a_{n,i,d})}
{\max(\sum_{d=1}^D q_{n,d}\omega_d,\delta)},
\label{eq:coordinate-loss}
\end{equation}
and the full blockwise action reconstruction objective is weighted by the inverse masking probability:
\begin{equation}
\mathcal L_{\mathrm{act}}
=\frac1N\sum_{n=1}^N
\frac{1}{\max(\sum_{i=1}^H v_{n,i},1)}
\sum_{i=1}^H v_{n,i}\,m_{n,i}\,\frac{\ell_{n,i}}{\max(t_{n,b(i)},\epsilon)}.
\label{eq:reconstruction-loss}
\end{equation}
Because $\mathbb E[m_{n,i}/t_{n,b(i)}\mid t]=1$, Equation~\ref{eq:reconstruction-loss} weights masked positions inversely by their selection rate across corruption levels and propagates gradients end-to-end through $D_\psi$, $F_\theta$, and $E_\phi$.

\subsection{Blockwise Decoding and Closed-Loop Execution}
\label{sec:method-decoding}

At inference, \method{} prefills the multimodal prefix $C$ into a key--value cache once and generates action blocks sequentially for $b=0,\ldots,M-1$. Within block $b$, action slots are initialized to the zero sentinel and updated over a preset proposal schedule using the shared backbone:
\begin{equation}
H_b^{(r)}=F_{\theta}\!\left(C,\,E_{\phi}(A_{\mathrm{final}}^{(<b)}),\,E_{\phi}(A^{(b,r)})\right)_b,\qquad \widehat A^{(b,r)}=D_{\psi}(H_b^{(r)}).
\end{equation}
To prioritize high-confidence actions during multi-step commitment, a lightweight token-risk head predicts the log reconstruction error $\log(\ell_{n,i}+10^{-6})$ from detached hidden states; proposals are ranked by negative predicted log-risk, and optional refinement rounds remask and re-predict low-scoring slots using the same backbone (Figure~\ref{fig:risk-guided-decoding}). Intermediate proposal rounds read from a read-only view of the prefix cache, and once block $b$ finishes, its final action vectors $A_{\mathrm{final}}^{(b)}$ are re-encoded and appended to the persistent KV cache for subsequent blocks. Every proposal and refinement step is executed by $F_\theta$ without an external action expert. Appendix~\ref{app:method-details} details the risk head, cache lifecycle, and inference hyperparameters.

\section{Experiments}
\label{sec:experiments}

We evaluate \method{} across three questions: (i) how a single shared diffusion VLM performs on standard closed-loop manipulation benchmarks (Section~\ref{sec:main-results}); (ii) how blockwise partition topology, pretraining coordinate weighting, and shared-space objective design govern closed-loop control (Sections~\ref{sec:ablation-topology}--\ref{sec:ablation-objective}); and (iii) where trajectory representation, temporal coupling, and parameter capacity reside across VLA families under our Action Interface Audit (Section~\ref{sec:temporal-response}). Appendix~\ref{app:experiment-details} provides complete training recipes, inference configurations, and additional evaluations on decoding budgets, replanning frequency, and representation stress tests.

\subsection{Evaluation protocol and scope}
\label{sec:experiment-protocol}

Our main benchmark evaluation covers regular LIBERO, LIBERO-Plus, and CALVIN ABC--D~\citep{libero,liberoplus,calvin}. Regular LIBERO evaluates 50 episodes per task across four suites (LIBERO-Spatial, LIBERO-Object, LIBERO-Goal, and LIBERO-10; 2,000 episodes total); LIBERO-Plus evaluates all 10,030 perturbed tasks across the four suites; CALVIN ABC--D evaluates all 1,000 official five-task instruction chains in the unseen environment D after training on environments A, B, and C. We report success rate (\%) on LIBERO and LIBERO-Plus, and average completed chain length (AvgLen) alongside task-$k$ chain success rates (SR@$k$) on CALVIN.

Multi-embodiment pretraining (60,000 updates followed by a 30,000-update continuation on the 80.2M-frame mixture) and CALVIN finetuning (5,000 updates) use eight GPUs with global batch size 1,536 (192 per GPU), while regular-LIBERO finetuning uses eight GPUs with global batch size 448 (56 per GPU). Appendix~\ref{app:training-recipe} documents the dataset mixture, training stages, and inference hyperparameters.

\subsection{Main benchmark results}
\label{sec:main-results}

Table~\ref{tab:main-libero} compares \method{} against representative discrete, action-expert, and diffusion-based VLA policies across regular LIBERO, LIBERO-Plus, and CALVIN ABC--D. On regular LIBERO, \method{} achieves $99.0\%$ on Spatial, $100.0\%$ on Object, $98.8\%$ on Goal, and $94.8\%$ on LIBERO-10, reaching an overall average of $\mathbf{98.15\%}$ across 2,000 episodes---the highest average among all compared methods, outperforming continuous action-expert policies such as $\pi_0$ ($94.2\%$), $\pi_{0.5}$ ($97.5\%$), and X-VLA ($98.1\%$), as well as discrete and hybrid diffusion policies including Discrete Diffusion VLA ($96.4\%$) and Dream-VLA ($97.2\%$). On the 10,030-task LIBERO-Plus benchmark, \method{} attains the highest score on every suite---$87.7\%$ on Spatial (2,106/2,402), $88.0\%$ on Object (2,216/2,518), $83.0\%$ on Goal (2,150/2,591), and $85.1\%$ on LIBERO-10 (2,143/2,519)---yielding $\mathbf{85.9\%}$ (8,615/10,030) overall and surpassing EFFVLA ($79.8\%$) by $+6.1$ percentage points and OpenVLA-OFT ($69.6\%$) by $+16.3$ points. On the zero-shot environment-transfer benchmark CALVIN ABC--D, \method{} reaches an average completed chain length of $4.142$ across 1,000 chains, achieving $98.6\%$ SR@1 and $93.8\%$ SR@2 (second-best overall and higher than X-VLA's $97.1\%$ SR@1 and $92.6\%$ SR@2) while outperforming the shared-backbone diffusion policy LLaDA-VLA ($4.01$ AvgLen, $95.6\%$ SR@1) across all chain lengths.

\begin{table*}[t]
\centering
\vspace{-0.6em}
\scriptsize
\setlength{\tabcolsep}{1.2pt}
\renewcommand{\arraystretch}{1.08}
\resizebox{\textwidth}{!}{%
\begin{tabular}{@{}>{\raggedright\arraybackslash}p{1.62in} *{16}{r}@{}}
\toprule
& \multicolumn{5}{c}{Regular LIBERO (\%)} & \multicolumn{5}{c}{LIBERO-Plus (\%)} & \multicolumn{6}{c}{CALVIN ABC--D} \\
\cmidrule(lr){2-6}\cmidrule(lr){7-11}\cmidrule(lr){12-17}
Method / checkpoint & Spatial & Object & Goal & LIBERO-10 & Avg. & Spatial & Object & Goal & LIBERO-10 & Overall & AvgLen & SR@1 & SR@2 & SR@3 & SR@4 & SR@5 \\
\midrule
\multicolumn{17}{@{}l}{\textit{Discrete \& Compressed-Token VLAs}} \\
Octo~\citep{octo} & 78.9 & 85.7 & 84.6 & 51.1 & 75.1 & -- & -- & -- & -- & -- & -- & -- & -- & -- & -- & -- \\
OpenVLA~\citep{openvla} & 84.7 & 88.4 & 79.2 & 53.7 & 76.5 & 19.4 & 14.0 & 15.1 & 14.3 & 15.6 & -- & -- & -- & -- & -- & -- \\
$\pi_0$-Fast~\citep{pi0fast} & 96.4 & 96.8 & 88.6 & 60.2 & 85.5 & 74.4 & 72.7 & 57.5 & 43.4 & 61.6 & -- & -- & -- & -- & -- & -- \\
NORA~\citep{nora} & -- & -- & -- & -- & -- & 47.6 & 34.4 & 38.8 & 36.3 & 39.0 & -- & -- & -- & -- & -- & -- \\
WorldVLA (512$^2$)~\citep{worldvla} & 87.6 & 96.2 & 83.4 & 60.0 & 81.8 & 32.5 & 28.6 & 31.8 & 8.2 & 25.0 & -- & -- & -- & -- & -- & -- \\
UniVLA~\citep{univla} & 95.4 & 98.8 & 93.6 & 94.0 & 95.5 & 55.5 & 36.7 & 40.7 & 39.9 & 42.9 & \underline{4.41} & \textbf{98.9} & \textbf{94.8} & \textbf{89.0} & \underline{82.8} & \underline{75.1} \\
\midrule
\multicolumn{17}{@{}l}{\textit{External Action-Expert \& Continuous-Head VLAs}} \\
$\pi_0$~\citep{pi0} & 96.8 & 98.8 & 95.8 & 85.2 & 94.2 & 60.7 & 61.4 & 44.9 & 48.4 & 53.6 & -- & -- & -- & -- & -- & -- \\
$\pi_{0.5}$~\citep{pi05} & 97.2 & 99.0 & 97.8 & \underline{96.0} & 97.5 & -- & -- & -- & -- & -- & -- & -- & -- & -- & -- & -- \\
SmolVLA 2.25B~\citep{smolvla} & 93.0 & 94.0 & 91.0 & 77.0 & 88.75 & -- & -- & -- & -- & -- & -- & -- & -- & -- & -- & -- \\
X-VLA~\citep{xvla} & 98.2 & 98.6 & 97.8 & \textbf{97.6} & \underline{98.1} & -- & -- & -- & -- & -- & \textbf{4.43} & 97.1 & 92.6 & \underline{88.5} & \textbf{84.4} & \textbf{78.8} \\
OpenVLA-OFT~\citep{openvlaoft} & 97.6 & 98.4 & 97.9 & 94.5 & 97.1 & 84.0 & 66.5 & 63.0 & 66.4 & 69.6 & -- & -- & -- & -- & -- & -- \\
OpenVLA-OFT$_w$~\citep{openvlaoft} & -- & -- & -- & -- & -- & 62.5 & 56.0 & 53.3 & 52.2 & 55.8 & -- & -- & -- & -- & -- & -- \\
OpenVLA-OFT$_m$~\citep{openvlaoft} & -- & -- & -- & -- & -- & 75.4 & 77.1 & 56.2 & 63.9 & 67.9 & -- & -- & -- & -- & -- & -- \\
RIPT-VLA~\citep{ript} & \underline{98.6} & 98.6 & \textbf{99.0} & 93.8 & 97.5 & \underline{85.8} & 64.3 & 58.0 & 67.5 & 68.4 & -- & -- & -- & -- & -- & -- \\
IMLE-VLA~\citep{imle} & 98.0 & \underline{99.8} & 98.2 & \underline{96.0} & 98.0 & -- & -- & -- & -- & -- & -- & -- & -- & -- & -- & -- \\
EFFVLA~\citep{effvla} & -- & -- & -- & -- & -- & 84.4 & \underline{84.5} & \underline{75.5} & \underline{74.7} & \underline{79.8} & -- & -- & -- & -- & -- & -- \\
\midrule
\multicolumn{17}{@{}l}{\textit{Shared-Backbone VLAs \& Embodied VLM (\evlm)}} \\
LLaDA-VLA~\citep{lladavla} & -- & -- & -- & -- & -- & -- & -- & -- & -- & -- & 4.01 & 95.6 & 87.8 & 79.5 & 73.9 & 64.5 \\
Discrete Diffusion VLA~\citep{ddvla} & 97.2 & 99.4 & 96.8 & 92.2 & 96.4 & -- & -- & -- & -- & -- & -- & -- & -- & -- & -- & -- \\
Dream-VLA~\citep{dreamvla} & 97.6 & 98.8 & 97.2 & 95.0 & 97.2 & -- & -- & -- & -- & -- & -- & -- & -- & -- & -- & -- \\
\textbf{\method{} (\evlm, ours)} & \textbf{99.0} & \textbf{100.0} & \underline{98.8} & 94.8 & \textbf{98.15} & \textbf{87.7} & \textbf{88.0} & \textbf{83.0} & \textbf{85.1} & \textbf{85.9} & 4.142 & \underline{98.6} & \underline{93.8} & 83.2 & 73.7 & 64.9 \\
\bottomrule
\end{tabular}%
}
\vspace{-0.4em}
\caption{Main results across regular LIBERO, LIBERO-Plus, and CALVIN ABC--D, grouped by architectural paradigm. Regular-LIBERO Avg.\ is the arithmetic mean of its four suite scores. Bold and underline mark the best and second-best result per column. The $\pi_{0.5}$ and IMLE-VLA rows follow the comparison reported by Hosseinkhani et al.~\citep{imle}. EFFVLA reports suite-level LIBERO-Plus averages, while \method{} and the remaining LIBERO-Plus rows report the official 10,030-task benchmark with Long labeled LIBERO-10. CALVIN reports average completed chain length and sequence success rates (SR@$1$--$5$) on ABC$\rightarrow$D.}
\vspace{-0.6em}
\label{tab:main-libero}
\end{table*}

\subsection{Blockwise Horizon Partition, Pretraining Weighting, and Execution Horizon}
\label{sec:ablation-topology}

\begin{figure*}[t]
\centering
\vspace{-0.4em}
\includegraphics[width=0.98\textwidth]{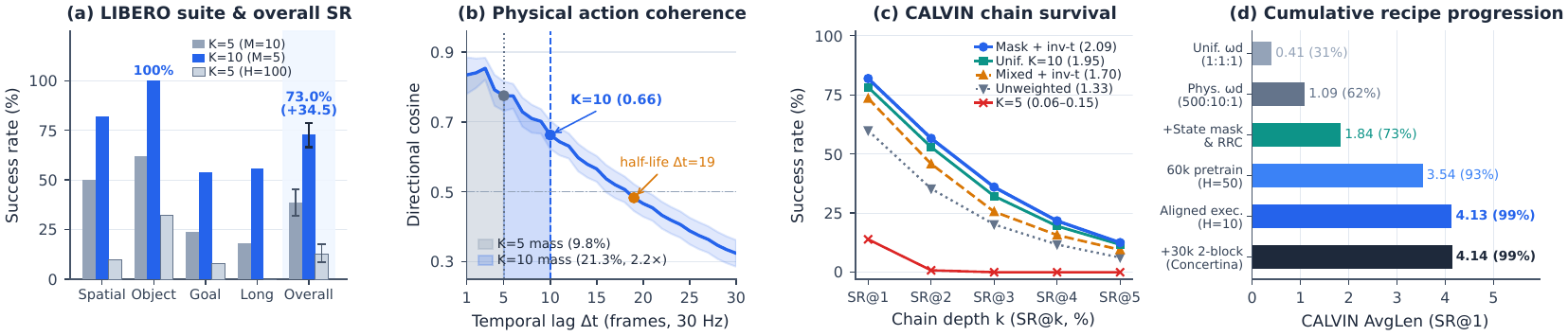}
\vspace{-0.8em}
\caption{\textbf{Blockwise horizon partition, physical action coherence, and cumulative recipe progression.} (a) Regular-LIBERO suite-level and overall success over 200 episodes (40 tasks $\times$ 5 episodes, Wilson 95\% CIs) at fixed 2{,}400-step finetuning budgets, showing consistent gains across all four suites ($38.5\%\rightarrow 73.0\%$ overall). (b) Directional cosine similarity of physical action increments across temporal lags $\Delta t=1\dots 30$ over 999{,}830 AgiBot frames (30\,Hz, 95\% task-bootstrap CI); enlarging the block size from $K=5$ to $K=10$ more than doubles intra-block coherent action mass ($9.8\%\rightarrow 21.3\%$, $2.17\times$). (c) Official 1{,}000-sequence CALVIN ABC--D chain survival curves (SR@$1$--$5$) across block partitions ($K=5$ vs.\ $K=10$) and pretraining corruption/weighting schedules ($1.333\rightarrow 1.701\rightarrow 2.087$ AvgLen). (d) Cumulative CALVIN ABC--D progression from uniform coordinate weights ($0.41$) to physical weighting, state masking, 60k multi-block pretraining, aligned single-block execution ($4.135$), and two-block continued pretraining ($4.142$).}
\vspace{-0.4em}
\label{fig:ablation-topology}
\end{figure*}

\paragraph{Block partition and physical action coherence.}
To examine how intra-block bidirectional context and inter-block causality affect policy learning, we first conduct a controlled ablation at a fixed horizon of $H=50$, varying the partition from ten blocks of five ($M=10, K=5$) to five blocks of ten ($M=5, K=10$) under identical pretraining and downstream finetuning budgets (2{,}400 finetuning updates on LIBERO and 2{,}600 on CALVIN), and evaluating on 200 regular-LIBERO episodes (40 tasks $\times$ 5 episodes) and all 1{,}000 official CALVIN ABC--D sequences. Enlarging each block from $K=5$ to $K=10$ substantially improves closed-loop control across every suite and chain depth (Figure~\ref{fig:ablation-topology}a,c). On regular LIBERO, the five-blocks-of-ten ($M=5, K=10$) layout reaches $73.0\%$ (146/200) compared with $38.5\%$ (77/200) for ten blocks of five ($M=10, K=5$)—raising Spatial from $50\%$ to $82\%$, Object from $62\%$ to $100\%$, Goal from $24\%$ to $54\%$, and LIBERO-10 from $18\%$ to $56\%$ ($+34.5$ percentage points overall, with 32 of 40 tasks improving, seven tying, and only one declining). On CALVIN ABC--D, average completed chain length rises $35\times$ from $0.055$ to $1.945$, SR@1 jumps from $5.1\%$ to $78.2\%$, and SR@5 rises from $0\%$ to $11.7\%$. Replacing uniform visible substitution with Gaussian corruption at $M=10, K=5$ leaves LIBERO success unchanged at $38.5\%$ (CALVIN AvgLen $0.147$), while extending the horizon to $H=100$ with twenty blocks of five ($M=20, K=5$) reduces valid unpadded trajectory starts ($12.5\%$ on LIBERO, $0.115$ on CALVIN). Moreover, holding the $M=5, K=10$ partition fixed on CALVIN (Figure~\ref{fig:ablation-topology}c), adding inverse-$t$ loss weighting (Equation~\ref{eq:reconstruction-loss}) lifts three-state mixed corruption from $1.333$ to $1.701$ AvgLen ($+13.6\text{ pp}$ SR@1), and switching to mask-only inverse-$t$ reconstruction further raises performance to $\mathbf{2.087}$ AvgLen ($81.9\%$ SR@1, $12.5\%$ SR@5).

To understand why block width $K=10$ is critical, we analyze the temporal autocorrelation of physical action increments (translation deltas, $\mathrm{SO}(3)$ geodesic rotation deltas, and gripper changes) across 999{,}830 frames from 680 pretraining episodes spanning all 34 AgiBot tasks at 30\,Hz (Figure~\ref{fig:ablation-topology}b). Directional cosine similarity remains $\mathbf{0.775}$ at lag $\Delta t=5$ (95\% task-bootstrap CI $[0.732,0.816]$), $\mathbf{0.663}$ at lag $\Delta t=10$ ($[0.623,0.703]$), and first drops below $0.5$ at lag $\Delta t=19$ (and below $1/e$ at lag $\Delta t=27$). Because contiguous manipulation primitives (such as reaching, aligning, and grasping) remain strongly directional over $15$--$20$ frames, a narrow $K=5$ partition under random block phasing retains only $9.8\%$ of positive lag-$1\dots 30$ coherence mass within the same bidirectional block, splitting ongoing motion phases across causal block boundaries. Doubling the block size to $K=10$ captures $21.3\%$ of coherent trajectory mass within each block ($2.17\times$), enabling contiguous control steps to be reconstructed jointly as a coherent motion unit.

\begin{table*}[t]
\centering
\vspace{-0.2em}
\scriptsize
\setlength{\tabcolsep}{4.2pt}
\renewcommand{\arraystretch}{1.08}
\resizebox{\textwidth}{!}{%
\begin{tabular}{@{}l c c c c r r r r r r@{}}
\toprule
Pretraining configuration & Coord.\ weights $\omega_d$ & State mask & Early-layer freeze & Visual RRC & AvgLen & SR@1 & SR@2 & SR@3 & SR@4 & SR@5 \\
\midrule
\textbf{Full pretraining recipe} & $500:10:1$ & \cmark & Layers 0--1 & \cmark & \textbf{1.837} & \textbf{73.1} & \textbf{50.4} & \textbf{30.5} & \textbf{18.9} & \textbf{10.8} \\
Full trainable backbone, w/o state mask & $500:10:1$ & \xmark & None (all 36L) & \cmark & \underline{1.589} & \underline{69.9} & \underline{42.7} & \underline{23.9} & \underline{14.4} & \underline{8.0} \\
w/o proprioceptive state mask & $500:10:1$ & \xmark & Layers 0--1 & \cmark & 1.093 & 61.9 & 30.1 & 11.2 & 4.4 & 1.7 \\
w/o state mask, w/o visual RRC & $500:10:1$ & \xmark & Layers 0--1 & \xmark & 1.027 & 56.4 & 26.7 & 12.3 & 4.9 & 2.4 \\
Full trainable backbone, w/o state mask \& RRC & $500:10:1$ & \xmark & None (all 36L) & \xmark & 0.607 & 40.6 & 15.2 & 3.6 & 1.0 & 0.3 \\
Uniform coordinate weights, w/o state mask & $1:1:1$ & \xmark & Layers 0--1 & \cmark & 0.535 & 36.5 & 13.0 & 3.0 & 0.8 & 0.2 \\
Uniform coordinate weights, w/ state mask & $1:1:1$ & \cmark & Layers 0--1 & \cmark & 0.406 & 31.0 & 7.5 & 1.5 & 0.6 & 0.0 \\
\bottomrule
\end{tabular}%
}
\vspace{-0.4em}
\caption{Controlled pretraining ablation evaluating physical coordinate weighting ($\omega_d$), proprioceptive state masking, early-layer freezing, and random-resized crop (RRC). Every configuration runs 4{,}800 pretraining updates on AgiBot followed by 2{,}600 CALVIN finetuning updates (global batch 1{,}536) and is evaluated on all 1{,}000 official CALVIN ABC--D chains.}
\vspace{-0.4em}
\label{tab:ablation-pretrain}
\end{table*}

\paragraph{Pretraining coordinate weighting and state masking.}
Because translation deltas in physical meters have orders-of-magnitude smaller numerical variance than normalized rotation and gripper coordinates, unscaled regression under-weights translational accuracy inside the shared backbone. Table~\ref{tab:ablation-pretrain} evaluates a controlled 7-configuration pretraining study (4{,}800 AgiBot pretraining steps $\rightarrow$ 2{,}600 CALVIN finetuning steps, 1{,}000 evaluation chains per configuration) isolating our physical coordinate weighting $\omega_d=(500,10,1)$ in Equation~\ref{eq:coordinate-loss}, proprioceptive state masking, and visual random-resized crop (RRC). Three matched findings stand out: (i) replacing uniform $1:1:1$ coordinate weights with $500:10:1$ raises CALVIN AvgLen by $\mathbf{+1.431}$ when state masking is active ($1.837$ vs.\ $0.406$, SR@1 $73.1\%$ vs.\ $31.0\%$) and by $\mathbf{+0.558}$ without state masking ($1.093$ vs.\ $0.535$); (ii) masking proprioceptive state during pretraining encourages the backbone to ground trajectories in visual-language context rather than relying solely on state extrapolation, lifting AvgLen by $\mathbf{+0.744}$ under $500:10:1$ weighting ($1.837$ vs.\ $1.093$, $+11.2\text{ pp}$ SR@1 and $+9.1\text{ pp}$ SR@5); and (iii) enabling visual RRC enhances spatial generalization when all 36 backbone layers are trained end-to-end, improving AvgLen by $\mathbf{+0.982}$ ($1.589$ vs.\ $0.607$).

\begin{table*}[t]
\centering
\vspace{-0.2em}
\scriptsize
\setlength{\tabcolsep}{4.0pt}
\renewcommand{\arraystretch}{1.08}
\resizebox{\textwidth}{!}{%
\begin{tabular}{@{}l l c c r r r r r r@{}}
\toprule
Training \& execution configuration & Pretraining stage & Downstream $(H,M,K)$ & Exec.\ $H_{\mathrm{exec}}$ & AvgLen & SR@1 & SR@2 & SR@3 & SR@4 & SR@5 \\
\midrule
Multi-block finetuning ($H=50$) & 60k ($H=50, M=5, K=10$) & $(50, 5, 10)$ & $10$ & 3.535 & 92.6 & 82.5 & 70.2 & 59.3 & 48.9 \\
Two-block finetuning ($H=20$), 2-block exec. & 60k ($H=50, M=5, K=10$) & $(20, 2, 10)$ & $20$ & 3.768 & 95.3 & 85.8 & 74.8 & 64.9 & 56.0 \\
Two-block finetuning ($H=20$), 1-block exec. & 60k ($H=50, M=5, K=10$) & $(20, 2, 10)$ & $10$ & 3.924 & 96.4 & 89.2 & 79.9 & 69.2 & 57.7 \\
Aligned 1-block finetuning ($H=10$) & 60k ($H=50, M=5, K=10$) & $(10, 1, 10)$ & $10$ & \underline{4.135} & \textbf{98.7} & \textbf{93.9} & \textbf{84.6} & 72.1 & 64.2 \\
\midrule
Single-block +40k cont.\ (trainable interfaces) & 60k + 40k ($H=10, M=1, K=10$) & $(10, 1, 10)$ & $10$ & 3.786 & 97.6 & 89.0 & 75.5 & 64.3 & 52.2 \\
Single-block +40k cont.\ (frozen interfaces) & 60k + 40k ($H=10, M=1, K=10$) & $(10, 1, 10)$ & $10$ & 3.934 & 97.9 & 90.3 & 79.7 & 67.2 & 58.3 \\
\textbf{Multi-block +30k cont.\ (\method{})} & 60k + 30k ($H=20, M=2, K=10$) & $(10, 1, 10)$ & $10$ & \textbf{4.142} & \underline{98.6} & \underline{93.8} & \underline{83.2} & \textbf{73.7} & \textbf{64.9} \\
\bottomrule
\end{tabular}%
}
\vspace{-0.4em}
\caption{Synergy between multi-block pretraining and receding-horizon closed-loop execution on CALVIN ABC--D (1{,}000 chains per row, 5{,}000 finetuning updates). Aligning the downstream finetuning horizon to the $K=10$ receding-horizon execution block raises AvgLen from $3.535$ to $4.135$, while retaining multi-block causal-bidirectional context during continued pretraining ($H=20, M=2, K=10$) outperforms single-block continued pretraining ($H=10, M=1, K=10$) by $+0.208$ AvgLen (95\% paired bootstrap CI $[+0.119, +0.298]$).}
\vspace{-0.4em}
\label{tab:ablation-horizon-progression}
\end{table*}

\paragraph{Multi-block pretraining and receding-horizon closed-loop execution.}
Multi-block packed training ($M>1, K=10$) and receding-horizon closed-loop control play complementary roles across pretraining and deployment. During pretraining, supervising multiple consecutive blocks in a single packed forward pass requires the shared VLM backbone to model cross-block phase transitions and longer-horizon task progression. During closed-loop control, the robot executes one coherent $K=10$ action block ($H_{\mathrm{exec}}=10$, corresponding to $0.33\text{\,s}$ at $30\text{\,Hz}$) before incorporating fresh visual observations. As shown in Table~\ref{tab:ablation-horizon-progression}, aligning the downstream finetuning horizon directly with this executed $K=10$ block concentrates gradient updates on the active control segment, raising CALVIN AvgLen from $3.535$ ($H=50$) to $3.768$ ($H=20, H_{\mathrm{exec}}=20$), $3.924$ ($H=20, H_{\mathrm{exec}}=10$), and $\mathbf{4.135}$ at $H=10, H_{\mathrm{exec}}=10$ ($98.7\%$ SR@1, $64.2\%$ SR@5). At the same time, multi-block context remains vital during pretraining: continuing pretraining for 30{,}000 updates with two-block $H=20\ (M=2, K=10)$ reaches $\mathbf{4.142}$ AvgLen ($64.9\%$ SR@5), outperforming a 40{,}000-update single-block $H=10\ (M=1, K=10)$ continuation ($3.786$ and $3.934$) by $\mathbf{+0.208}$ AvgLen (95\% paired bootstrap CI $[+0.119, +0.298]$, $+6.6\text{ pp}$ SR@5). Multi-block pretraining thus endows the shared backbone with multi-phase dynamical representations that directly strengthen single-block closed-loop execution.

\subsection{Co-Design of Action Representation and Objective in the Shared Hidden Space}
\label{sec:ablation-objective}

\begin{figure*}[t]
\centering
\vspace{-0.4em}
\includegraphics[width=0.98\textwidth]{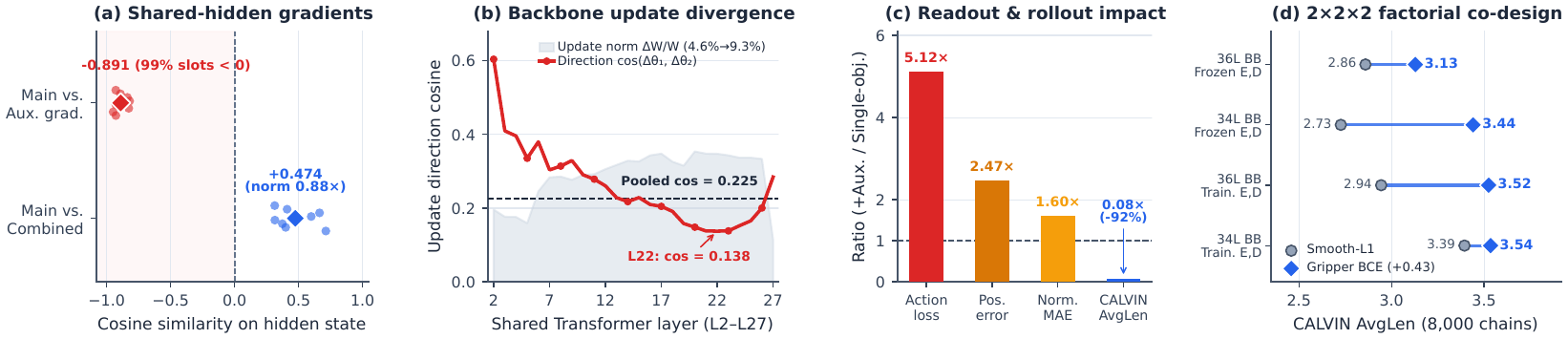}
\vspace{-0.8em}
\caption{\textbf{Co-design of action representation and supervision objective in the shared hidden space.} (a) Shared-hidden gradient cosine between the main relative-action objective and auxiliary absolute-action objective across eight fixed windows is $-0.891$ ($99\%$ of action slots $<0$), reducing combined-vs-main alignment to $+0.474$ (auxiliary/main norm ratio $0.88\times$). (b) Layerwise parameter update divergence across all 26 active Transformer layers (L2--L27): while relative update norm $\Delta W/W$ rises from $4.6\%$ to $9.3\%$ with identical overall magnitude ($1.003\times$), the parameter update direction cosine plunges from $0.604$ at Layer 2 to $0.138$ at Layer 22 (pooled cosine $0.225$). (c) Resulting readout degradation ($5.12\times$ teacher action loss, $2.47\times$ position error, $1.60\times$ normalized MAE) and $92.0\%$ closed-loop CALVIN collapse ($2.970\rightarrow 0.237$). (d) Complete $2\times 2\times 2$ factorial co-design across 8{,}000 CALVIN chains on 4.46B BARD-VL-B4, showing consistent gains from probability-valued gripper BCE ($+0.427$ AvgLen) and trainable action interfaces ($+0.311$).}
\vspace{-0.4em}
\label{fig:ablation-objective}
\end{figure*}

\paragraph{Counterfactual dual-objective intervention.}
Enabling a VLM to natively output continuous actions without degrading its pretrained representations requires care: when manipulation and cognition share a single hidden space, that space is not a passive feature buffer onto which incompatible coordinate readouts can be attached in parallel. We test this principle via a matched counterfactual intervention on the 8.35B shared-backbone checkpoint ($H=25$) that holds every model and optimization setting identical except for the supervision target: the single-objective baseline supervises relative actions alone, whereas the dual-objective variant attaches a frozen auxiliary decoder supervising absolute actions at loss weight $0.5$, with all 21/21 main action-interface tensors initialized bitwise identically from pretraining.

Attaching the second coordinate target induces severe gradient conflict directly on the shared backbone hidden states (Figure~\ref{fig:ablation-objective}a--c). Across eight fixed CALVIN windows, the pooled cosine similarity between the relative main gradient and absolute auxiliary gradient on the shared hidden state is $-0.891$, with $99\%$ of individual action-token gradients pointing in opposing directions. Even at weight $0.5$, the auxiliary gradient norm is $0.881\times$ the main gradient norm, reducing the cosine alignment between the combined gradient and the main objective to $0.474$. Inspecting the resulting parameter updates across all 26 trainable Transformer layers (Figure~\ref{fig:ablation-objective}b) reveals that while both runs make updates of virtually identical magnitude (relative $L_2$ update norm rising from $4.6\%$ in shallow layers to $9.3\%$ in layers 17--22; overall norm ratio $1.003\times$), the parameter update direction cosine drops from $0.604$ at Layer 2 down to $\mathbf{0.138}$ at Layer 22 (pooled cosine $0.225$)—precisely the middle and late layers where trajectory coupling resides. Consequently, teacher-forced main loss increases $5.12\times$, one-pass position error increases $2.47\times$, normalized MAE increases $1.60\times$, and closed-loop CALVIN chain length drops by $92.0\%$ from $2.970$ to $0.237$ across 1{,}000 sequences (Figure~\ref{fig:ablation-objective}c). In a unified model, the action coordinate representation and its supervision objective jointly define the optimization geometry of the shared VLM backbone and must be co-designed as a single coherent interface.

\paragraph{Factorial ablation of coordinate objective, interface co-adaptation, and end-to-end backbone training.}
To quantify how coordinate loss design and parameter co-adaptation interact on the 4.46B BARD-VL-B4 architecture, we evaluate a complete $2\times 2\times 2$ factorial design (5{,}000 updates per arm, 8{,}000 official CALVIN ABC--D evaluation chains in total; Figure~\ref{fig:ablation-objective}d and Table~\ref{tab:ablation-calvin-factorial}). Three consistent conclusions emerge across all matched pairs: (i) supervising the gripper closure coordinate with probability-valued binary cross-entropy (Appendix~\ref{app:method-details}) rather than Smooth-$L_1$ yields a mean gain of $\mathbf{+0.427}$ AvgLen ($3.408$ vs.\ $2.981$) and $\mathbf{+11.2\text{ pp}}$ SR@5 ($45.95\%$ vs.\ $34.73\%$), improving all four matched pairs ($+0.715$, $+0.141$, $+0.270$, $+0.581$) by calibrating grasp closure probabilities across multi-stage chains; (ii) co-adapting the position-wise action interfaces $(E_\phi, D_\psi)$ with the backbone adds $\mathbf{+0.311}$ mean AvgLen ($3.350$ vs.\ $3.039$) over frozen interfaces; and (iii) when the action interfaces and gripper objective are properly co-designed, training all 36 shared Transformer layers end-to-end ($3.525$ AvgLen, $92.7\%$ SR@1) and freezing only the first two stem layers ($3.535$ AvgLen, $92.6\%$ SR@1) both preserve backbone stability and achieve strong multi-step performance without requiring Knowledge Insulation or gradient blocking.

\begin{table*}[t]
\centering
\vspace{-0.2em}
\scriptsize
\setlength{\tabcolsep}{4.5pt}
\renewcommand{\arraystretch}{1.08}
\resizebox{\textwidth}{!}{%
\begin{tabular}{@{}l l l r r r r r r@{}}
\toprule
Gripper coordinate objective & Backbone training & Action interfaces ($E_\phi, D_\psi$) & AvgLen & SR@1 & SR@2 & SR@3 & SR@4 & SR@5 \\
\midrule
Probability-valued Gripper BCE & Freeze layers 0--1 (34L trainable) & Trainable & \textbf{3.535} & \underline{92.6} & \underline{82.5} & \underline{70.2} & \textbf{59.3} & \textbf{48.9} \\
Probability-valued Gripper BCE & Full trainable backbone (all 36L) & Trainable & \underline{3.525} & \textbf{92.7} & \textbf{82.9} & \textbf{71.1} & \underline{58.3} & \underline{47.5} \\
Probability-valued Gripper BCE & Freeze layers 0--1 (34L trainable) & Frozen & 3.441 & 91.1 & 80.8 & 68.7 & 57.0 & 46.5 \\
Smooth-$L_1$ on all coordinates & Freeze layers 0--1 (34L trainable) & Trainable & 3.394 & 91.2 & 81.6 & 67.4 & 55.1 & 44.1 \\
Probability-valued Gripper BCE & Full trainable backbone (all 36L) & Frozen & 3.129 & 87.0 & 74.5 & 60.5 & 50.0 & 40.9 \\
Smooth-$L_1$ on all coordinates & Full trainable backbone (all 36L) & Trainable & 2.944 & 87.7 & 71.6 & 57.3 & 43.8 & 34.0 \\
Smooth-$L_1$ on all coordinates & Full trainable backbone (all 36L) & Frozen & 2.859 & 85.9 & 71.1 & 54.7 & 42.8 & 31.4 \\
Smooth-$L_1$ on all coordinates & Freeze layers 0--1 (34L trainable) & Frozen & 2.726 & 84.9 & 68.3 & 51.5 & 38.5 & 29.4 \\
\midrule
\multicolumn{3}{@{}l}{\textit{Factorial main effect: Gripper Closure BCE vs.\ Smooth-$L_1$ (mean over 4 matched pairs)}} & \textbf{$+$0.427} & $+$4.2 & $+$6.5 & $+$9.9 & $+$11.1 & \textbf{$+$11.2} \\
\multicolumn{3}{@{}l}{\textit{Factorial main effect: Trainable vs.\ Frozen interfaces (mean over 4 matched pairs)}} & \textbf{$+$0.311} & $+$3.8 & $+$6.0 & $+$7.7 & $+$7.1 & $+$6.5 \\
\multicolumn{3}{@{}l}{\textit{Factorial main effect: Freeze VLM layers 0--1 vs.\ Full trainable BB (mean over 4 matched pairs)}} & \textbf{$+$0.160} & $+$1.6 & $+$2.5 & $+$2.6 & $+$3.8 & $+$4.1 \\
\bottomrule
\end{tabular}%
}
\vspace{-0.4em}
\caption{Full $2\times 2\times 2$ factorial ablation on 4.46B BARD-VL-B4 (initialized from the 60k pretraining checkpoint at $H=50, M=5, K=10$; 5{,}000 CALVIN updates at batch size 1{,}536; evaluated on all 1{,}000 official CALVIN ABC--D chains per configuration, 8{,}000 chains total).}
\vspace{-0.4em}
\label{tab:ablation-calvin-factorial}
\end{table*}

\subsection{Action Interface Audit: Measuring Manipulation--Cognition Coupling}
\label{sec:temporal-response}

To evaluate how conventional VLA paradigms and our \evlm{} (\method) couple manipulation with cognition across representation, computation, and parameters, we conduct the \textbf{Action Interface Audit} using three checkpoint-level metrics summarized in Table~\ref{tab:app-temporal} and Figure~\ref{fig:followup-diagnostics} (full protocols and sensitivity analyses in Appendices~\ref{app:representation-audit}--\ref{app:temporal-response}):

\begin{table*}[t]
\centering
\vspace{-0.2em}
\scriptsize
\setlength{\tabcolsep}{3.2pt}
\resizebox{\textwidth}{!}{%
\begin{tabular}{@{}l l l r r r r r l@{}}
\toprule
Checkpoint & Action representation & Trajectory generator & $H$ & Valid $D$ & Interface NLR & Full NLR & ASCR & Trajectory coupling locus \\
\midrule
$\pi_0$ base~\citep{pi0} & 32D continuous & External flow expert & 50 & 32 & 0 & 8.141\% & 16.51\% & External action expert \\
$\pi_{0.5}$ base~\citep{pi05} & 32D continuous & External flow expert & 50 & 32 & 0 & 7.141\% & 19.17\% & External action expert \\
SmolVLA base~\citep{smolvla} & 32D continuous & External flow expert & 50 & 32 & -- & -- & 22.19\% & External action expert \\
X-VLA base~\citep{xvla} & 20D continuous & External 24L Transformer & 50 & 20 & -- & -- & 35.10\% & External 24L Transformer \\
\midrule
\method{} (\evlm) pretraining & 20D canonical latent & Shared VLM backbone (36L) & 50 & 20 & 0 & 4.058\% & \textbf{0.21\%} & Shared VLM backbone \\
\method{} (\evlm) CALVIN & 10D absolute latent & Shared VLM backbone (36L) & 50 & 10 & 0 & 4.103\% & \textbf{0.21\%} & Shared VLM backbone \\
\method{} (\evlm) LIBERO & 10D relative latent & Shared VLM backbone (36L) & 50 & 10 & 0 & 1.217\% & \textbf{0.21\%} & Shared VLM backbone \\
\bottomrule
\end{tabular}%
}
\vspace{-0.4em}
\caption{Action Interface Audit: Trajectory Coupling Attribution (TCA) and Action-Specific Capacity Ratio (ASCR) across external-expert VLAs and our \evlm{} (\method). All three \method{} rows evaluate 4.46B BARD-VL-B4 checkpoints at $H=50$ under deterministic float32 math-only SDPA ($\epsilon=0.02$). Under packed blockwise BD3 attention ($M=5, K=10$), the three \method{} checkpoints likewise yield zero interface NLR and $2.595\%$, $3.163\%$, and $1.078\%$ full-path NLR, respectively.}
\vspace{-0.4em}
\label{tab:app-temporal}
\end{table*}

\paragraph{(i) Error-Geometry Consistency (EGC) and Latent Manifold Rank: Representational Coupling.}
Fixing a unit training-event error (one token substitution for discrete tokenizers; a unit coordinate error for continuous heads), we measure the resulting physical action error distribution on 512 contiguous CALVIN windows ($H=30, D=7$) and define $\mathrm{EGC}=p_{95}/p_{05}$. While continuous coordinate supervision guarantees $\mathrm{EGC}=1$ by construction, 256-bin uniform quantization~\citep{openvla} yields $p_{05}=0.0549$ and $p_{95}=1.3804$, an error-geometry span of $\mathrm{EGC}=25.14\times$ ($25.57\times$ on an expanded 8{,}192-window sample) because cross-entropy penalizes adjacent and distant bin errors identically. For compressed frequency-domain tokenizers, single-token substitutions break sequence decodability in $82.23\%$ of windows for FAST+~\citep{pi0fast} and $77.93\%$ for repaired MolmoAct2-FAST, and shape-valid edits perturb a median of $30/30$ timesteps across the entire window (Figure~\ref{fig:followup-diagnostics}a). Conversely, auditing the local Jacobian $\partial E_\phi(a)/\partial a\in\mathbb R^{2560\times 20}$ of \method's continuous action encoder across pretraining checkpoints confirms that $E_\phi$ embeds actions as a smooth, non-collapsed 20D chart inside the VLM hidden space (Figure~\ref{fig:followup-diagnostics}b): its entropy effective rank evolves from $19.78/20$ at initialization to $19.30$, $18.90$, and $18.63/20$ at $5\text{k}$, $10\text{k}$, and $15\text{k}$ updates (median condition number $1.73\rightarrow 3.22$), while retaining a $0.855$--$0.931$ pairwise distance Spearman correlation with physical coordinates.

\begin{figure*}[t]
\centering
\vspace{-0.4em}
\includegraphics[width=0.98\textwidth]{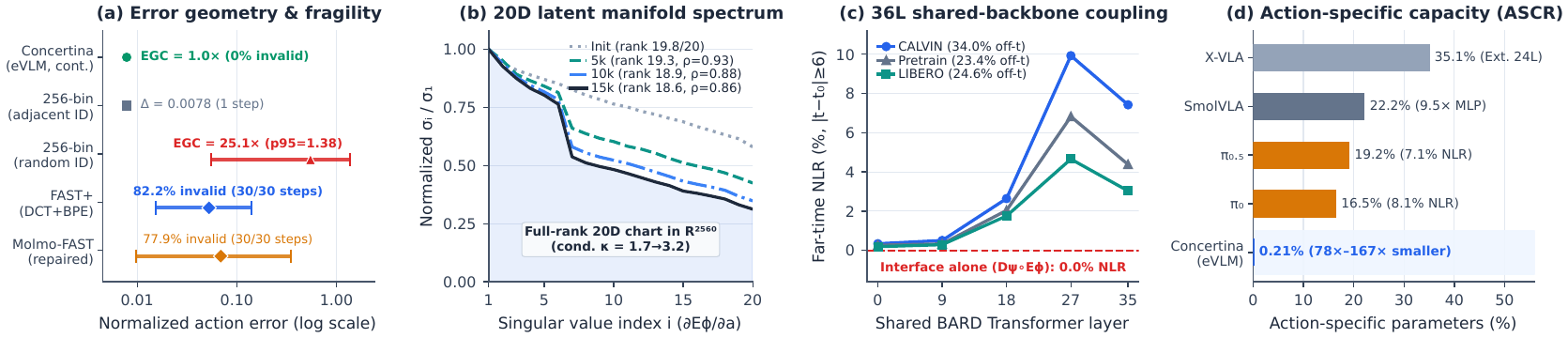}
\vspace{-0.8em}
\caption{\textbf{Action Interface Audit across representational, computational, and parameter coupling.} (a) Physical action error span ($p_{05}$--$p_{50}$--$p_{95}$) and structural decoding fragility under single-token substitutions: continuous coordinates (\method{}) preserve unit error geometry ($\mathrm{EGC}=1.0\times$, $0\%$ invalid), whereas 256-bin quantization induces a $25.1\times$ error span and compressed DCT+BPE tokenizers (FAST+, MolmoAct2-FAST) suffer $82.2\%$ and $77.9\%$ invalid trajectory shapes while perturbing $30/30$ steps on valid decodes. (b) Normalized Jacobian singular value spectrum ($\sigma_i/\sigma_1$) of \method's continuous action encoder $\partial E_\phi(a)/\partial a\in\mathbb R^{2560\times 20}$ across pretraining updates, maintaining a full-rank 20D manifold (effective rank $18.63/20$, Spearman $\rho=0.86$--$0.93$). (c) Layerwise emergence of cross-time trajectory coupling ($\mathrm{NLR}$, $|t-t_0|\geq 6$) across the 36-layer BARD-VL-B4 backbone versus $0.0\%$ for the position-wise interface ($D_\psi\circ E_\phi$) alone. (d) Action-Specific Capacity Ratio ($\mathrm{ASCR}$) across VLA families, showing that \method{} uses $78\times$--$167\times$ fewer action-specific parameters ($0.21\%$) than external-expert architectures ($16.5\%$--$35.1\%$).}
\vspace{-0.4em}
\label{fig:followup-diagnostics}
\end{figure*}

\paragraph{(ii) Trajectory Coupling Attribution (TCA): Computational Coupling.}
To localize where cross-time trajectory dependencies are computed, we perturb a single input action coordinate $(t_0,d_0)$ by $\pm\epsilon$ and measure the output action response tensor $R\in\mathbb R^{H\times D}$ via central finite differences under deterministic float32 math-only SDPA (zero repeated-forward jitter). We define the \emph{non-local response ratio} as the fraction of response energy propagating at least six timesteps away:
\begin{equation}
\mathrm{NLR}=\frac{\sum_{|t-t_0|\geq 6}\|R_{t,:}\|_2^2}{\|R\|_F^2}.
\end{equation}
Comparing the position-wise action interface alone ($D_\psi\circ E_\phi$) against the full shared-backbone forward pass ($D_\psi\circ F_\theta\circ E_\phi$) across three 4.46B BARD-VL-B4 \method{} checkpoints ($H=50$, 36 layers) shows that the interface path has strictly zero cross-time response ($\mathrm{NLR}=0$), whereas including a single forward pass of the shared VLM backbone produces full-path $\mathrm{NLR}$ of $4.058\%$ (60k pretraining), $4.103\%$ (CALVIN downstream), and $1.217\%$ (LIBERO downstream), with total off-time energy ($t\neq t_0$) reaching $23.4\%$--$34.0\%$ (Table~\ref{tab:app-temporal}). Layerwise hidden-state hooks across the 36-layer BARD backbone reveal that cross-time coupling is near zero in shallow layers ($0.19\%$--$0.34\%$ at layer 0 and $0.29\%$--$0.50\%$ at layer 9) and forms progressively in the middle and late Transformer layers, peaking at $4.66\%$--$9.93\%$ at layer 27 and $3.03\%$--$7.42\%$ at layer 35 (Figure~\ref{fig:followup-diagnostics}c). By contrast, auditing $\pi_0$, $\pi_{0.5}$, SmolVLA, and X-VLA~\citep{pi0,pi05,smolvla,xvla} shows that their trajectory coupling and coordinate subspace alignment occur inside external modules outside the VLM: in $\pi_0$ and $\pi_{0.5}$, full-path $\mathrm{NLR}$ ($8.141\%$ and $7.141\%$) is generated entirely inside the external flow expert; in $\pi_0$ and SmolVLA, 32D input/output principal-subspace overlap is near random chance before the expert's action-time MLP ($0.99\times$ and $0.94\times$) and rises to $4.36\times$ and $9.51\times$ only after the expert MLP; and in X-VLA, skipping the external 24-layer action Transformer retains only $3.72\%$ of same-dimension action energy.

\paragraph{(iii) Action-Specific Capacity Ratio (ASCR): Parameter Coupling.}
Finally, quantifying the parameter share dedicated exclusively to action generation (Table~\ref{tab:app-temporal} and Figure~\ref{fig:followup-diagnostics}d) shows that external-expert VLAs allocate $16.51\%$ ($\pi_0$), $19.17\%$ ($\pi_{0.5}$), $22.19\%$ (SmolVLA~\citep{smolvla}), and $35.10\%$ (X-VLA~\citep{xvla}) of total model capacity to separate action generators and bridges. In \method's 4.46B BARD-VL-B4 architecture, there is no separate action Transformer ($0\%$ external generator capacity), and the active 4-layer action encoder and decoder projections contain only $9.49\text{M}$ parameters ($\mathrm{ASCR}=0.21\%$, a $78\times$--$167\times$ reduction in action-specific parameter share). Together, EGC, TCA, and ASCR confirm that \method{} preserves physical error geometry while unifying trajectory computation and parameter capacity inside the shared VLM backbone.

\section{Conclusion}
\label{sec:conclusion}
We presented \method, an \emph{Embodied Vision-Language Model} (\evlm) built on the principle that manipulation and cognition should be embodied within the same model rather than decoupled across a bolted-on VLA pipeline---enabling a pretrained diffusion VLM to natively output precise continuous actions while preserving its visual-language generalization, model architecture, and native training paradigm. By representing each control timestep as a continuous latent action token inside a shared diffusion VLM backbone and supervising temporally ordered action blocks directly in physical coordinates via packed teacher forcing, \method{} eliminates both discrete action tokenizers and external action experts while retaining the backbone's native corrupt-and-reconstruct operator. Our Action Interface Audit across representative VLA families and \method{} demonstrates that an \evlm{} preserves unit error geometry ($\mathrm{EGC}=1$) and a full-rank 20D action manifold ($18.63/20$ effective rank), uses only $0.21\%$ active action-interface parameters ($\mathrm{ASCR}$), and forms cross-time trajectory coupling directly within the middle and late layers of the shared 36-layer VLM backbone ($\mathrm{NLR}=1.22\%$--$4.10\%$). Controlled ablations further establish why blockwise bidirectional horizon partitioning ($K=10$ capturing $2.17\times$ more coherent action mass and yielding $+34.5$ points on LIBERO), physical coordinate weighting with state masking ($+1.431$ and $+0.744$ CALVIN AvgLen), multi-block pretraining continuation ($+0.208$ AvgLen over single-block continuation), and unified single-objective supervision with probability-valued gripper BCE ($+0.427$ mean AvgLen across 8{,}000 factorial chains) are essential when manipulation and cognition share one hidden space. On standard closed-loop benchmarks, \method{} achieves $98.15\%$ average success on regular LIBERO, $85.9\%$ on the 10{,}030-task LIBERO-Plus benchmark, and $4.142$ average completed chain length on CALVIN ABC--D.

\bibliographystyle{iclr2027_conference}
\bibliography{references}

\clearpage
\appendix
\section{Implementation and Objective Details}
\label{app:method-details}

\subsection{Architecture and Coordinate Interface}
\label{app:method-config}

\begin{figure*}[ht]
\centering
\includegraphics[width=0.84\textwidth]{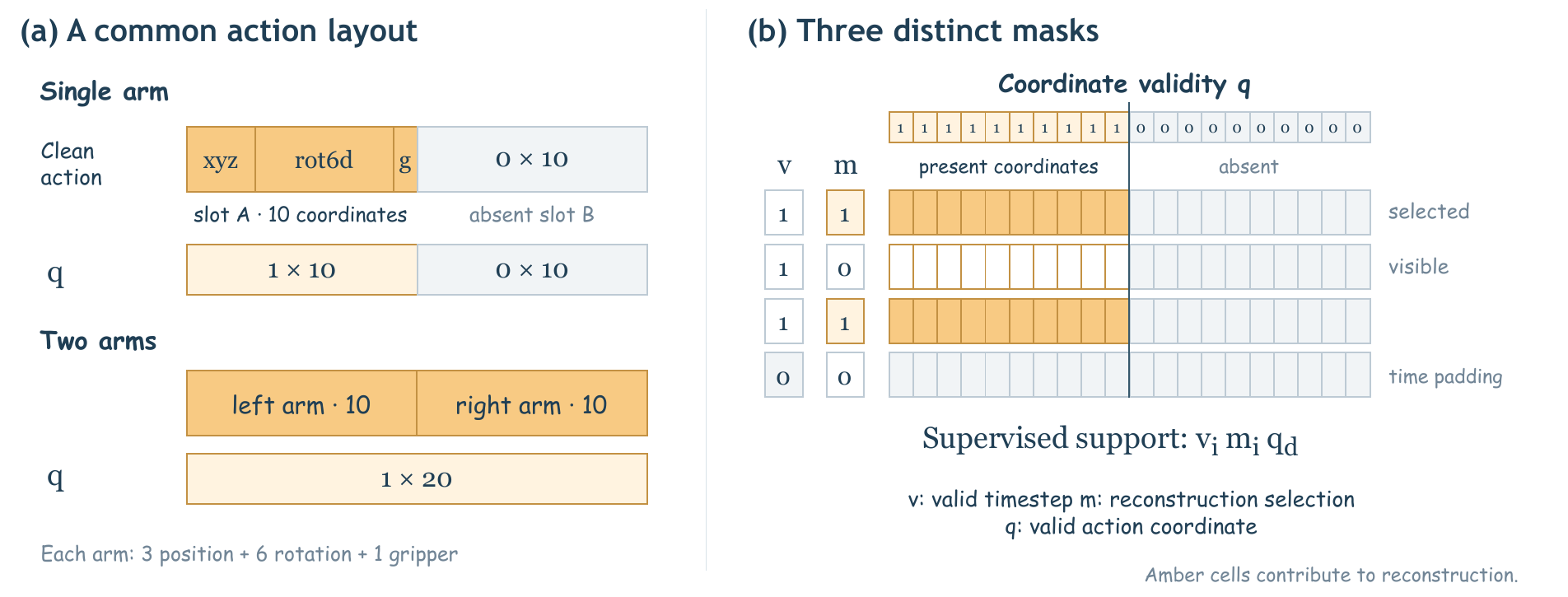}
\caption{\textbf{Validity masks separate embodiment coordinates, reconstruction selection, and time padding.} The canonical 20D layout provides fixed slots for single-arm and bimanual embodiments; coordinate validity $q$, masking indicator $m$, and timestep validity $v$ play orthogonal roles in the supervised objective.}
\label{fig:action-validity}
\end{figure*}

\method{} instantiates the shared diffusion VLM backbone $F_\theta$ from BARD-VL-B4-Mask-4B-Instruct~\citep{bard} (36 Transformer layers, hidden dimension $d=2560$) with a unified $D=20$ canonical coordinate width supporting both single-arm and bimanual embodiments. Each arm occupies ten canonical coordinates: three Cartesian translations, a continuous six-dimensional rotation representation, and one gripper closure coordinate. Inactive embodiment dimensions are zeroed at the input and masked from the loss via the coordinate-validity indicator $q_{n,d}\in\{0,1\}$ (Figure~\ref{fig:action-validity}).

Both the position-wise action encoder $E_\phi$ and action decoder $D_\psi$ are four-layer MLPs with hidden width 1{,}024, SiLU activations, and zero dropout. Proprioceptive robot state is encoded through the same input projection $E_\phi$ because it shares the canonical 20D coordinate layout. \method{} introduces no action-boundary special tokens, embodiment-specific adapters, or soft prompts. Coordinate reconstruction weights are set to $\omega_d=(500,500,500,10,10,10,10,10,10,1)$ per arm across both the numerator and denominator of Equation~\ref{eq:coordinate-loss}, specifying the relative weighting among translation, 6D rotation, and gripper coordinates. Continuous pose coordinates are supervised with Smooth-$L_1$ loss, and the gripper coordinate is supervised with probability-valued binary cross-entropy (Section~\ref{app:prob-gripper}).

\subsection{Masking Schedules, Padding, and Packed Attention}

In the standard mask-only corruption schedule (Section~\ref{sec:method-objective}), each block $b$ draws a stratified corruption rate $t_{n,b}\in[\epsilon,1]$ with $\epsilon=10^{-3}$, and each timestep $i$ in block $b$ draws an independent binary mask indicator $m_{n,i}\sim\operatorname{Bernoulli}(t_{n,b})$ replacing the entire vector $a_{n,i}$ with the zero sentinel before encoding through $E_\phi$. Stage-1 pretraining also supports the three-state mixed corruption schedule inherited from BARD~\citep{bard}, which assigns each position a clean, visible-substitution, or masked state with pre-flooring probabilities
\begin{equation}
\begin{split}
p_{\mathrm{clean}}(t)&=1-\sin(\pi t/2),\\
p_{\mathrm{sub}}(t)&=\sin(\pi t/2)+\cos(\pi t/2)-1,\\
p_{\mathrm{mask}}(t)&=1-\cos(\pi t/2),
\end{split}
\label{eq:mixed-corruption}
\end{equation}
each floored at $10^{-3}$ and renormalized to sum to one, where visible substitution replaces the vector with a random vector and supervises both substituted and masked positions. For mask-only corruption, conditioning on the block corruption levels $t$ yields
\begin{equation}
\mathbb E_m\!\left[\frac{m_i}{t_{b(i)}}\ell_i(m)\,\middle|\,t\right]
=\mathbb E_m[\ell_i(m)\mid m_i=1,t],
\end{equation}
so the inverse-$t$ weighting in Equation~\ref{eq:reconstruction-loss} weights each masked slot inversely by its selection probability across corruption levels.

During packed teacher-forced training, invalid timesteps ($v_{n,i}=0$) are masked out as attention keys and excluded from supervision; to prevent all-masked attention rows in Flash/SDPA kernels, an invalid query attends only to itself. Within the packed sequence $X_{\mathrm{train}}=[C,\widetilde Z,Z]$, the multimodal prefix $C$ is causal and isolated from both action branches. Queries in the corrupted branch $\widetilde Z$ attend to $C$, their own corrupted block $\widetilde Z^{(b)}$, and strictly earlier clean blocks $Z^{(<b)}$, while queries in the clean branch $Z$ attend to $C$ and clean blocks $Z^{(\leq b)}$. Because both action branches share aligned positional encodings and predictions are read exclusively from $\widetilde Z$, the clean branch provides exact teacher-forced block context without leaking current-block targets.

\subsection{Probability-Valued Gripper Supervision}
\label{app:prob-gripper}

To unify heterogeneous gripper conventions across embodiments (e.g., $[-1,1]$ in CALVIN and LIBERO versus $[0,1]$ in AgiBot and DROID), we map raw gripper commands to a canonical closure coordinate $p=(g-g_{\mathrm{open}})/(g_{\mathrm{closed}}-g_{\mathrm{open}})\in[0,1]$. The action decoder outputs a scalar logit $r$ for the gripper channel, supervised via binary cross-entropy with target-entropy subtraction:
\begin{equation}
\rho_{\mathrm{grip}}(r,p)
=\operatorname{BCEWithLogits}(r,p)-\mathcal H(p)
=\operatorname{KL}\!\left(\operatorname{Ber}(p)\,\Vert\,
\operatorname{Ber}(\sigma(r))\right),
\end{equation}
where $\mathcal H(p)$ denotes the Bernoulli entropy of the target. Subtracting $\mathcal H(p)$ leaves the parameter gradient unchanged while setting the minimum loss to zero for both hard binary targets and continuous intermediate closure targets. At inference, predicted logits are mapped through the sigmoid function $\widehat p=\sigma(r)$ before commitment and re-encoding through $E_\phi$, ensuring that iterative refinement operates consistently in probability coordinates.

\subsection{Risk-Guided Decoding and KV-Cache Lifecycle}

\begin{figure*}[ht]
\centering
\includegraphics[width=0.84\textwidth]{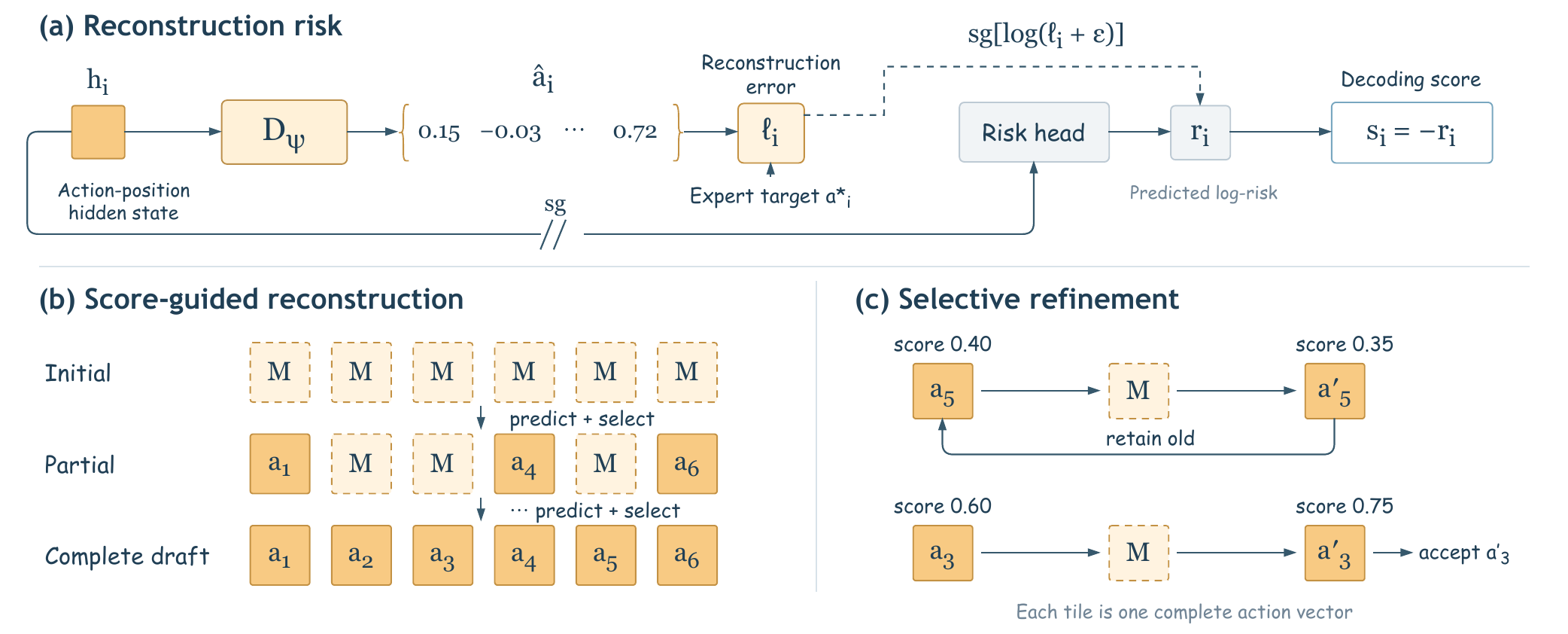}
\caption{\textbf{Risk-guided decoding scores candidate action tokens and supports selective iterative refinement.} The token-risk head predicts log reconstruction error from detached hidden states; candidate slots are committed in order of confidence, and optional refinement rounds remask and update low-scoring positions when the predicted score improves.}
\label{fig:risk-guided-decoding}
\end{figure*}

The token-risk head applies LayerNorm followed by a two-layer MLP (hidden width 256, SiLU activation, scalar output) to detached action hidden states $h_i$, predicting the log coordinate error $\log(\ell_{n,i}+10^{-6})$ under a Smooth-$L_1$ auxiliary loss weighted by $0.05$. Because its input is detached, the risk head provides a lightweight confidence score $s_i=-r_i$ for inference-time slot selection without perturbing backbone gradients during training.

During blockwise decoding of block $b$, pending action slots are initialized to the zero sentinel and committed over at most $D$ proposal rounds according to their confidence scores $s_i$, followed by up to $R$ selective refinement steps that remask low-scoring positions and accept updated predictions when confidence improves (Figure~\ref{fig:risk-guided-decoding}). Throughout proposal and refinement rounds within block $b$, the multimodal prefix cache and earlier block caches $B_{<b}$ remain read-only (as illustrated in Figure~\ref{fig:overview}, bottom right). Once block $b$ is finalized, its committed continuous actions are re-encoded through $E_\phi$ and appended to the KV cache for block $b+1$.

\section{Additional Experimental Details}
\label{app:experiment-details}

\subsection{Pretraining Mixture and Training Stages}
\label{app:training-recipe}

Tables~\ref{tab:pretraining-mixture}--\ref{tab:shared-training-settings} summarize the 80.2M-frame multi-embodiment pretraining mixture, the two-stage pretraining and downstream finetuning schedule, and the shared optimization and inference hyperparameters.

\begin{table}[ht]
\centering
\scriptsize
\setlength{\tabcolsep}{4.5pt}
\begin{tabular}{@{}lrrrr@{}}
\toprule
Dataset source & Subsets / views & Episodes & Frames & Sampling weight \\
\midrule
AgiBotWorld--Alpha & 34 / 34 & 32{,}735 & 46{,}156{,}708 & $44.44\%$ \\
DROID (30\,Hz aligned) & 1 / 2 & 97{,}685 & 27{,}348{,}488 & $33.33\%$ \\
RoboMIND Franka & 1 / 1 & 5{,}333 & 943{,}842 & $11.11\%$ \\
RoboMIND AgileX & 1 / 1 & 8{,}906 & 5{,}387{,}380 & $7.78\%$ \\
RoboMIND FR3 & 1 / 1 & 1{,}774 & 375{,}807 & $3.33\%$ \\
\midrule
\textbf{Total pretraining mixture} & \textbf{38 / 39} & \textbf{146{,}433} & \textbf{80{,}212{,}225} & \textbf{100.00\%} \\
\bottomrule
\end{tabular}
\caption{Multi-embodiment pretraining dataset mixture totaling 146{,}433 episodes and 80.2M frames across five robot manipulation sources.}
\label{tab:pretraining-mixture}
\end{table}

\begin{table}[ht]
\centering
\scriptsize
\setlength{\tabcolsep}{4pt}
\begin{tabularx}{\linewidth}{@{}l l l X@{}}
\toprule
Stage & Dataset & Updates & Configuration \\
\midrule
Stage-1 Pretraining & Multi-embodiment mixture (Table~\ref{tab:pretraining-mixture}) & 60{,}000 & Canonical 20D actions ($H=50, M=5, K=10$); active-dimension validity masks; 30\,Hz temporal alignment \\
Stage-2 Continued Pretraining & Multi-embodiment mixture (Table~\ref{tab:pretraining-mixture}) & 30{,}000 & Two-block continuation ($H=20, M=2, K=10$); mask-only inverse-$t$ objective; probability-valued gripper BCE \\
Downstream Finetuning & CALVIN ABC--D & 5{,}000 & Absolute end-effector actions; single-block horizon ($H=10, M=1, K=10$) \\
Downstream Finetuning & Regular LIBERO / LIBERO-Plus & 5{,}000 & Relative end-effector actions; single-block horizon ($H=10, M=1, K=10$) \\
\bottomrule
\end{tabularx}
\caption{Overview of pretraining and downstream finetuning stages.}
\label{tab:training-stages}
\end{table}

\begin{table}[ht]
\centering
\scriptsize
\setlength{\tabcolsep}{4pt}
\begin{tabularx}{\linewidth}{@{}l X@{}}
\toprule
Hyperparameter & Value \\
\midrule
Compute \& batch size & 8 GPUs; global batch size 1{,}536 (192/GPU) for pretraining and CALVIN finetuning; global batch size 448 (56/GPU) for LIBERO finetuning \\
Closed-loop execution & $H=10, K=10, M=1$; proposal budget $D=4$; refinement steps $R=5$; executed horizon $H_{\mathrm{exec}}=10$ \\
Warmup \& checkpointing & 80-step linear warmup; checkpoints saved every 1{,}250 updates; seed 42 \\
\bottomrule
\end{tabularx}
\caption{Shared optimization and closed-loop inference configuration.}
\label{tab:shared-training-settings}
\end{table}

\subsection{Multi-Embodiment Data Curation and Boundary Handling}
\label{app:pretraining-cleaning}

To ensure high trajectory quality across heterogeneous robot datasets, we apply an offline multi-stage curation pipeline prior to pretraining. The pipeline verifies timestamp continuity, finite coordinate ranges, and frame-to-action alignment across all 146{,}433 episodes. Non-informative static intervals are identified by cross-checking canonical end-effector velocity, joint motion channels, and visual optical flow, while semantic phase rules explicitly preserve functional holds such as gripper stabilization and pouring phases. Whenever a non-informative static segment is excised from a demonstration, the curation pipeline resets local episode boundaries so that each retained motion segment forms a contiguous trajectory, and windows crossing a segment boundary are padded using the explicit timestep-validity mask $v_{n,i}\in\{0,1\}$.

\subsection{Decoding Budget and Closed-Loop Replanning Frequency}
\label{app:focused-refinement}

Because \method{} generates an entire $K=10$ continuous action block within the shared VLM backbone, it supports both single-pass parallel decoding ($D=1, R=0$: all 10 action tokens predicted in one backbone forward pass after prefix prefill) and multi-step risk-guided proposal and refinement ($D\leq 4, R=0$ and $D\leq 4, R=5$). We evaluate the effect of decoding budget and closed-loop replanning frequency across three complementary protocols: (i) an offline open-loop reconstruction probe over 32 fixed trajectory windows ($2{,}880$ non-gripper coordinates per arm, $10{,}000$ paired bootstrap resamples) on both CALVIN and LIBERO; (ii) a 100-sequence closed-loop CALVIN ABC--D evaluation (sequences $0$--$99$); and (iii) a 50-episode closed-loop LIBERO-Spatial evaluation ($10\text{ tasks}\times 5\text{ initial states}$) comparing \method{} against $\pi_{0.5}$ across flow steps and execution horizons $E\in\{1,10,50\}$.

\begin{table*}[ht]
\centering
\scriptsize
\setlength{\tabcolsep}{4.0pt}
\renewcommand{\arraystretch}{1.08}
\resizebox{\textwidth}{!}{%
\begin{tabular}{@{}l c r r r r r r r@{}}
\toprule
& & & \multicolumn{2}{c}{Offline 32-Window MAE} & \multicolumn{4}{c}{CALVIN Closed-Loop (100 chains)} \\
\cmidrule(lr){4-5}\cmidrule(lr){6-9}
Decoding strategy & Mean BB passes & Latency $p_{50}$ (ms) & CALVIN MAE & LIBERO MAE & AvgLen & SR@1 (\%) & SR@3 (\%) & SR@5 (\%) \\
\midrule
$D=1, R=0$ (single-pass parallel) & $1.00$ & $65.8$ & $0.06282$ & $0.27692$ & $3.92$ & $95.0$ & $80.0$ & \textbf{59.0} \\
$D\leq 4, R=0$ (proposals only) & $2.00$ & $91.7$ & $0.06279$ & $0.27326$ & $3.90$ & $93.0$ & $79.0$ & $56.0$ \\
$D\leq 4, R=5$ (with risk refinement) & $7.00$ & $223.0$ & \textbf{0.06196} & \textbf{0.26852} & \textbf{4.04} & \textbf{97.0} & \textbf{82.0} & \textbf{59.0} \\
\bottomrule
\end{tabular}%
}
\caption{Offline reconstruction accuracy (32 windows, $2{,}880$ non-gripper coordinates per arm) and 100-sequence closed-loop CALVIN ABC--D performance across decoding budgets at fixed checkpoint identity ($H=10, K=10, M=1$). Risk-guided refinement ($D\leq 4, R=5$) reduces offline MAE on 25/32 CALVIN windows ($\Delta=-0.00084$, 95\% paired bootstrap CI $[-0.00125,-0.00043]$) and 25/32 LIBERO windows ($\Delta=-0.00840$ vs.\ $D=1, R=0$, 95\% CI $[-0.01316,-0.00400]$), while single-pass parallel decoding ($D=1, R=0$) already achieves $3.92$ AvgLen at $65.8\text{\,ms}$ median latency ($6.58\text{\,ms}$ per executed control step).}
\label{tab:focused-refinement}
\end{table*}

\begin{table}[ht]
\centering
\scriptsize
\setlength{\tabcolsep}{4.0pt}
\renewcommand{\arraystretch}{1.08}
\resizebox{\linewidth}{!}{%
\begin{tabular}{@{}l l c c r r@{}}
\toprule
Method & Decoding budget & Chunk $H$ & Exec.\ $E$ & Success (50 eps) & Mean capped steps \\
\midrule
\method{} & $D=1, R=0$ (1 pass) & $10$ & $10$ & \textbf{50/50 (100.0\%)} & \textbf{102.40} \\
\method{} & $D\leq 4, R=5$ (7 passes) & $10$ & $10$ & \textbf{50/50 (100.0\%)} & $102.84$ \\
\method{} & $D\leq 4, R=5$ (replan every step) & $10$ & $1$ & 46/50 (92.0\%) & $118.82$ \\
\midrule
$\pi_{0.5}$~\citep{pi05} & Flow 2 steps & $50$ & $10$ & 49/50 (98.0\%) & $105.92$ \\
$\pi_{0.5}$~\citep{pi05} & Flow 5 steps & $50$ & $10$ & 49/50 (98.0\%) & $106.08$ \\
$\pi_{0.5}$~\citep{pi05} & Flow 10 steps (default) & $50$ & $10$ & 48/50 (96.0\%) & $115.26$ \\
$\pi_{0.5}$~\citep{pi05} & Flow 10 steps (no replan) & $50$ & $50$ & 42/50 (84.0\%) & $133.60$ \\
\bottomrule
\end{tabular}%
}
\caption{Closed-loop decoding budget and replanning frequency on LIBERO-Spatial ($10\text{ tasks}\times 5\text{ initial states}=50\text{ episodes}$ per arm; failed episodes capped at 280 steps). Executing a full $E=10$ block before replanning outperforms step-by-step replanning ($E=1$, $+8.0\text{ pp}$) by preserving intra-block motion coherence, whereas open-loop 50-step execution in $\pi_{0.5}$ ($E=50$) degrades success by $-12.0\text{ pp}$ (95\% paired bootstrap CI $[-24.0, -2.0]$).}
\label{tab:app-replanning-libero}
\end{table}

Table~\ref{tab:focused-refinement} and Table~\ref{tab:app-replanning-libero} highlight three findings. First, single-pass parallel generation ($D=1, R=0$) is already highly accurate: because the shared 36-layer VLM backbone jointly contextualizes all 10 action slots within the block, a single forward pass achieves $100.0\%$ (50/50) success on the LIBERO-Spatial benchmark subset in $102.40$ mean steps and $3.92$ AvgLen on CALVIN at $65.8\text{\,ms}$ per 10-step block ($6.58\text{\,ms}$ per control step). Second, iterative risk-guided refinement ($D\leq 4, R=5$) consistently improves open-loop reconstruction on 25 of 32 windows on both CALVIN and LIBERO and raises closed-loop CALVIN AvgLen from $3.92$ to $4.04$. Third, executing the full $E=10$ coherent block outperforms both extremes of replanning frequency: replanning at every single control step ($E=1$) discards the committed trajectory structure and increases completion steps by $+15.98$ ($92.0\%$ success), while executing a 50-step chunk without visual feedback in $\pi_{0.5}$ ($E=50$) drops success from $96.0\%$ to $84.0\%$.

\subsection{Action Interface Audit: Error-Geometry Consistency (EGC) Protocol}
\label{app:representation-audit}

The representation audit evaluates 512 contiguous CALVIN relative-action windows ($H=30, D=7$, normalized range $[-1,1]$, seed 20260901), together with an expanded 8{,}192-window sample and four independently seeded 2{,}048-window samples (Tables~\ref{tab:app-representation}--\ref{tab:app-representation-followup}). The uniform quantizer uses 256 bin boundaries and 255 reconstruction centers (bin width $2/255$), evaluated over 10{,}000 random single-token substitutions; frequency-compressed tokenizers (FAST+~\citep{pi0fast} and MolmoAct2-FAST with byte-fallback repair for 44 out-of-vocabulary symbols) are evaluated under one single-token substitution per window.

\begin{table}[ht]
\centering
\scriptsize
\setlength{\tabcolsep}{3.5pt}
\resizebox{\linewidth}{!}{%
\begin{tabular}{@{}lrrrrr@{}}
\toprule
Representation & Unperturbed MAE & Mean tokens & EGC ($p_{95}/p_{05}$) & Invalid shape (\%) & Valid RMSE median \\
\midrule
Continuous coordinates (\method) & $0.0000$ & $30.0$ & $1.00$ & $0.00$ & \textup{n/a} \\
Uniform quantizer (256 bins) & $0.0023$ & $210.0$ & $25.14$ & $0.00$ & \textup{n/a} \\
FAST+~\citep{pi0fast} & $0.0200$ & $57.7$ & -- & $82.23$ & $0.0526$ \\
MolmoAct2-FAST (repaired) & $0.0200$ & $90.2$ & -- & $77.93$ & $0.0690$ \\
\bottomrule
\end{tabular}%
}
\caption{Error-Geometry Consistency (EGC) and representation diagnostics on 512 CALVIN windows ($H=30, D=7$). Unperturbed MAE measures round-trip reconstruction error; EGC, invalid shape, and valid RMSE refer to single-token substitution.}
\label{tab:app-representation}
\end{table}

\begin{table}[ht]
\centering
\scriptsize
\setlength{\tabcolsep}{2.5pt}
\begin{tabular}{@{}lrrrrrr@{}}
\toprule
Run & Windows & FAST+ invalid (\%) & FAST+ RMSE median & Molmo invalid (\%) & Molmo RMSE median & Time support \\
\midrule
Canonical & 512 & 82.23 & 0.0526 & 77.93 & 0.0690 & 30/30 \\
Expanded & 8{,}192 & 84.52 & 0.0590 & 79.24 & 0.1173 & 30/30 \\
Four-seed range & $4\times2{,}048$ & 81.30--85.11 & 0.0552--0.0631 & 76.22--79.00 & 0.0966--0.1171 & 30/30 \\
\bottomrule
\end{tabular}
\caption{Expanded representation stress test across 512, $4\times 2{,}048$, and 8{,}192 windows. Invalid shape is the fraction of single-token edits that fail to decode into $30\times 7$ coefficients. Time support counts affected steps with a coordinate change exceeding 0.01 normalized units.}
\label{tab:app-representation-followup}
\end{table}

\subsection{Action Interface Audit: Trajectory Coupling Attribution (TCA) and Capacity (ASCR)}
\label{app:temporal-response}

The Trajectory Coupling Attribution (TCA) probe compares the position-wise action interface composition ($D_\psi\circ E_\phi$) against one complete shared-backbone forward pass ($D_\psi\circ F_\theta\circ E_\phi$) on the 4.46B BARD-VL-B4 \method{} checkpoints ($H=50$, 36 Transformer layers). Central finite differences are evaluated at 11 time/dimension coordinates ($t_0\in\{0,12,25,37,49\}$ at $d_0=0$, plus $d_0\in\{1,\ldots,6\}$ at $t_0=25$) in deterministic float32 math-only SDPA with zero repeated-forward jitter. Table~\ref{tab:app-temporal-stability} verifies that the resulting non-local response ratios ($\mathrm{NLR}$, $|t-t_0|\geq 6$) remain invariant across perturbation magnitudes $\epsilon\in\{0.01,0.02,0.04\}$.

\begin{table}[ht]
\centering
\scriptsize
\setlength{\tabcolsep}{4.5pt}
\begin{tabular}{@{}lrrrr@{}}
\toprule
Checkpoint (4.46B BARD-VL-B4) & $\epsilon=0.01$ & $\epsilon=0.02$ & $\epsilon=0.04$ & Interface NLR \\
\midrule
\method{} pretraining ($H=50, D=20$) & $4.058\%$ & $4.058\%$ & $4.056\%$ & $0$ \\
\method{} downstream (CALVIN, $D=10$) & $4.103\%$ & $4.103\%$ & $4.102\%$ & $0$ \\
\method{} downstream (LIBERO, $D=10$) & $1.215\%$ & $1.217\%$ & $1.216\%$ & $0$ \\
\bottomrule
\end{tabular}
\caption{Full-path NLR stability across finite-difference step sizes $\epsilon\in\{0.01,0.02,0.04\}$ on the 4.46B BARD-VL-B4 checkpoints. Each cell averages 11 coordinate probes with zero repeated-forward jitter; interface-only NLR is identically zero.}
\label{tab:app-temporal-stability}
\end{table}

In the 4.46B BARD-VL-B4 \method{} architecture ($d=2560$), the active 4-layer action encoder $E_\phi$ ($4{,}744{,}704$ parameters) and 4-layer action decoder $D_\psi$ ($4{,}742{,}164$ parameters) total $9{,}486{,}868$ active interface parameters out of $4{,}447{,}963{,}669$ active model parameters ($\mathrm{ASCR}=0.213\%$, or $10{,}147{,}861$ parameters / $0.228\%$ including the $660{,}993$-parameter token-risk head), with zero separate action-generator layers.

\end{document}